\documentclass[journal]{IEEEtran}
\usepackage{cite}
\usepackage{amsmath,amsfonts}
\usepackage{algorithmic}
\usepackage{algorithm}
\usepackage{array}
\usepackage[caption=false,font=normalsize,labelfont=sf,textfont=sf]{subfig}
\usepackage{textcomp}
\usepackage{stfloats}
\usepackage{url}
\usepackage{verbatim}
\usepackage{graphicx}
\usepackage{placeins}
\usepackage{amsthm}
\usepackage{multirow}
\usepackage{graphicx}

\theoremstyle{plain}

\makeatletter
\def\th@plain{%
  \thm@notefont{}%
  \itshape
  \def\@begintheorem##1##2{%
    \item[\hskip\labelsep \theorem@headerfont ##1\ ##2.]}%
  \def\@opargbegintheorem##1##2##3{%
    \item[\hskip\labelsep \theorem@headerfont ##1\ ##2\ (##3).]}%
}
\def\theorem@headerfont{\normalfont\bfseries}
\makeatother

\begin{document}

\title{LI-DSN: A Layer-wise Interactive Dual-Stream Network for EEG Decoding}

\author{Chenghao Yue\textsuperscript{*}, 
    Zhiyuan Ma\textsuperscript{*}, 
    Zhongye Xia,
    Xinche Zhang,
    Yisi Zhang\textsuperscript{\dag},
    Xinke Shen\textsuperscript{\dag},
    Sen Song\textsuperscript{\dag}
    \thanks{This work was supported by research grants from the National Natural Science Foundation of China under Grant 2025ZD0215701, Grant T2341003, Grant U2336214, in part by the Beijing Natural Science Foundation under Grant L257019, in part by the Shenzhen Science and Technology Innovation Committee under Grant RCBS20231211090748082.}
    \thanks{*~These authors contributed equally to this work.}
    \thanks{\dag~Corresponding author.}
    \thanks{Zhiyuan Ma, Xinche Zhang and Sen Song are with the School of Biomedical Engineering, Tsinghua Laboratory of Brain and Intelligence, and IDG/McGovern Institute for Brain Research, Tsinghua University, Beijing 100084, China (e-mail: ma-zy25@mails.tsinghua.edu.cn; zhang-xc24@mails.tsinghua.edu.cn; songsen@tsinghua.edu.cn).}
    \thanks{Chenghao Yue is with the School of Life Sciences, Tsinghua University, Beijing 100084, China (e-mail: 2024261945@qq.com).}
    \thanks{Zhongye Xia and Xinke Shen are with the Department of Biomedical Engineering, Southern University of Science and Technology, Shenzhen 518055, China (e-mail: Xia\_Zhongye@outlook.com; shenxk@sustech.edu.cn).}
    \thanks{Yisi Zhang is with Department of Psychological and Cognitive Sciences, Tsinghua University, Beijing, 100084, China (e-mail: zhangyisi@tsinghua.edu.cn).}
}
    
\markboth{Journal of \LaTeX\ Class Files,~Vol.~14, No.~8, August~2021}%
{Shell \MakeLowercase{\textit{et al.}}: A Sample Article Using IEEEtran.cls for IEEE Journals}

\maketitle

\begin{abstract}

Electroencephalography (EEG) provides a non-invasive window into brain activity, offering high temporal resolution crucial for understanding and interacting with neural processes through brain-computer interfaces (BCIs).  Current dual-stream neural networks for EEG often process temporal and spatial features independently through parallel branches, delaying their integration until a final, late-stage fusion. This design inherently leads to an “information silo” problem, precluding intermediate cross-stream refinement and hindering spatial-temporal decompositions essential for full feature utilization. We propose LI-DSN, a layer-wise interactive dual-stream network that facilitates progressive, cross-stream communication at each layer, thereby overcoming the limitations of late-fusion paradigms. LI-DSN introduces a novel Temporal-Spatial Integration Attention (TSIA) mechanism, which constructs a Spatial Affinity Correlation Matrix (SACM) to capture inter-electrode spatial structural relationships and a Temporal Channel Aggregation Matrix (TCAM) to integrate cosine-gated temporal dynamics under spatial guidance. Furthermore, we employ an adaptive fusion strategy with learnable channel weights to optimize the integration of dual-stream features. Extensive experiments across eight diverse EEG datasets, encompassing motor imagery (MI) classification, emotion recognition, and steady-state visual evoked potentials (SSVEP), consistently demonstrate that LI-DSN significantly outperforms 13 state-of-the-art (SOTA) baseline models, showcasing its superior robustness and decoding performance.  The code will be publicized after acceptance.

\end{abstract}

\begin{IEEEkeywords}
Brain-Computer Interface, EEG Decoding, Dual-Stream Network, Layer-Wise Interaction.
\end{IEEEkeywords}

\section{Introduction}

Electroencephalography (EEG)-based brain-computer interfaces (BCIs) establish direct communication pathways between neural activity and external devices, with wide-ranging applications in medical rehabilitation, assistive technologies, and cognitive assessment \cite{wolpaw2007brain,moghimi2013review}. Despite significant advancements in signal processing and machine learning, robust EEG decoding remains a formidable challenge. This complexity stems from several inherent characteristics: low signal-to-noise ratios, substantial inter-subject variability, and the intricate spatiotemporal dynamics that underpin cortical neural activity \cite{murad2024unveiling}. Deep learning has emerged as a highly promising transformative paradigm for EEG decoding, offering end-to-end feature learning directly from raw signals. While early Convolutional Neural Networks (CNNs) like EEGNet \cite{lawhern2018eegnet} and DeepConvNet \cite{schirrmeister2017deep} effectively extracted localized spatiotemporal patterns, their inherently limited receptive fields restricted modeling long-range temporal dependencies. This led to the integration of Transformer architectures, leveraging self-attention for global temporal modeling in hybrid CNN-Transformer models \cite{song2023eeg,xie2022transformer}. However, these single-stream paradigms struggle to simultaneously capture diverse temporal dynamics and complex spatial distributions, as a unified representation often lacks the capacity to sufficiently resolve the distinct spatiotemporal properties of EEG signals.

Recognizing the distinct advantages and complementary roles of different architectural components, parallel dual-stream or multi-branch networks have garnered increasing attention as a promising alternative for EEG decoding. By explicitly decoupling and processing different aspects of EEG signals through dedicated branches, these models aim to better capture the multifaceted nature of brain activity. For instance, early multi-branch CNN-based architectures combined separate convolutional pathways for temporal and spatial filtering, such as the Inception-based designs \cite{wang2025dbconformer,salami2022eeg}. More recently, models have explored integrating attention mechanisms into dual-stream designs, where different branches independently employ self-attention modules to refine features before a final fusion stage \cite{gong2023astdf,amin2021attention}. While these dual-stream strategies have demonstrated superior performance compared to single-branch architectures, they often fall short in fully exploiting the rich, interconnected nature of EEG. Specifically, existing dual-stream networks share a fundamental limitation: streams operate largely independently until final fusion. Temporal and spatial branches typically process their respective features through multiple layers with minimal or no intermediate communication, deferring feature combination to the classification stage via simple concatenation or basic learned gating. This late-fusion strategy inherently creates an "information silo" problem, leading to two critical consequences that hinder optimal EEG decoding. First, delaying integration forces temporal and spatial streams to evolve in isolated representation spaces, thereby missing vital opportunities for mutual contextualization and early-layer cross-stream guidance. Second, this decoupled processing conflicts with neurophysiological evidence, which indicates that EEG signals arise from coupled spatiotemporal dynamics \cite{pfurtscheller2001motor,jiruska2013synchronization}, meaning that independent feature extraction fails to adequately capture these intrinsic interdependencies.

\begin{figure}
    \centering
    \includegraphics[width=1\linewidth]{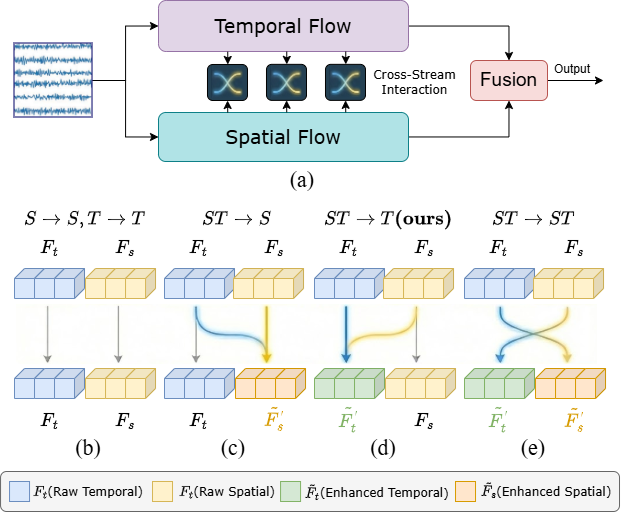}
    \caption{Exploration of dual-stream interaction strategies for EEG analysis, where $F_t$ and $F_s$ raw temporal and spatial features respectively, $\tilde{F'_t}$ and $\tilde{F'_s}$ denote enhanced features after interaction. (a) Dual-stream architecture with cross-flow interactions between temporal and spatial streams. (b) Baseline with no cross-stream interaction before fusion.  (c) Interaction outputs integrated only into the spatial stream. (d) Interaction outputs integrated only into the temporal stream. (e) Bidirectional interaction feeding both streams.}
    \label{fig:head}
\end{figure}

To overcome these limitations and unlock the full potential of coupled spatiotemporal modeling, we propose LI-DSN (Layer-wise Interactive Dual-Stream Network). LI-DSN introduces a novel paradigm enabling progressive cross-stream communication at each layer of the network. The core innovation is the Temporal-Spatial Integration Attention (TSIA) mechanism, specifically developed to allow temporal features to iteratively query and integrate relevant spatial context throughout the network's depth. Distinct from late-fusion strategies, TSIA dynamically computes two complementary matrices within each layer to capture coupled dynamics: (1) a Spatial Affinity Correlation Matrix (SACM), which incorporates electrode position embeddings to capture geometric inter-electrode relationships, and (2) a Temporal Channel Aggregation Matrix (TCAM), which utilizes a cosine-gated mechanism to discern how spatial patterns modulate oscillatory temporal dynamics. As illustrated in Figure \ref{fig:head}, we explored three distinct fusion paradigms and selected the asymmetric Spatiotemporal-to-Temporal (ST$\rightarrow$T) integration as our model's core strategy, ensuring that stable spatial priors effectively guide the refinement of temporal features (details are provided in Section \ref{sec:IM}). Finally, to optimize the decision-making process, we introduce an adaptive fusion strategy that employs learnable channel weights to capture data-driven electrode importance, alongside temporal attention pooling to emphasize discriminative temporal segments.

Our main contributions are summarized as follows:
\begin{itemize}
    \item We propose LI-DSN, a novel layer-wise interactive dual-stream network that enables progressive spatiotemporal integration through continuous interaction, effectively addressing the “information silo" problem inherent in late-fusion approaches.

    \item We design the TSIA mechanism to capture synergistic spatiotemporal representations by integrating spatial structural priors with temporal dynamics. To augment these representations, an adaptive fusion strategy is employed to selectively emphasize critical features through learnable weights.
    
    \item Comprehensive experiments on eight EEG datasets, spanning MI classification, emotion recognition, and SSVEP, consistently demonstrate LI-DSN's superior performance over 13 SOTA baselines.
\end{itemize}

\begin{table*}[!htbp]
    \caption{Summary of datasets and task types across different paradigms.}
    \renewcommand{\arraystretch}{1.2}
    \centering
    \footnotesize
    \setlength{\tabcolsep}{5pt}
    \begin{tabular}{clcccccc}
        \hline\hline
            Paradigm & Dataset & \begin{tabular}[c]{@{}c@{}}Number of\\Subjects\end{tabular} & \begin{tabular}[c]{@{}c@{}}Number of\\EEG Channels\end{tabular} & \begin{tabular}[c]{@{}c@{}}Sampling\\Rate (Hz)\end{tabular} & \begin{tabular}[c]{@{}c@{}}Trial Length\\(seconds)\end{tabular} & \begin{tabular}[c]{@{}c@{}}Number of\\Total Trials\end{tabular} & Task Types \\
        \hline
        \multirow{4}{*}{MI} 
        & BNCI2014001 & 9 & 22 & 250 & 4 & 1,296 & left/right hand \\
        & BNCI2014002 & 14 & 15 & 512 & 5 & 1,400 & right hand/both feet \\
        & BNCI2014004 & 9 & 3 & 250 & 5 & 1,080 & left/right hand \\
        & Zhou2016 & 4 & 14 & 250 & 5 & 409 & left/right hand \\
        \hline
        \multirow{2}{*}{Emotion Recognition}
        & SEED & 15 & 62 & 200 & 2 & 7,920 & negative/positive \\
        & FACED & 123 & 32 & 250 & 4 & 10,332 & negative/positive \\
        \hline
        \multirow{2}{*}{SSVEP} 
        & Nakanishi2015 & 9 & 8 & 256 & 4 & 1,620 & 12-target stimuli \\
        & Kalunga2016 & 12 & 8 & 256 & 2 & 945 & 4-target stimuli \\

        \hline\hline
    \end{tabular}
    \label{tab:datasets}
\end{table*}

\section{Related Work}

\subsection{CNN-Based EEG Decoding}
Convolutional Neural Networks (CNNs) have emerged as a dominant paradigm for EEG signal classification due to their effectiveness in extracting hierarchical features from raw data. Early seminal works include EEGNet \cite{lawhern2018eegnet}, which introduced compact architecture employing depthwise separable convolutions for spatial filtering and temporal convolutions for frequency-specific pattern extraction, and DeepConvNet/ShallowConvNet \cite{schirrmeister2017deep}, which explored depth-accuracy trade-offs, demonstrating that deeper architectures learn more abstract representations despite increased overfitting risk. Building upon these foundations, FBCNet \cite{mane2021fbcnet} explicitly models spectral information through parallel filter banks tuned to distinct frequency bands, mimicking classical Filter Bank Common Spatial Patterns within an end-to-end framework. Recent advancements have focused on augmenting CNN capabilities for complex spatiotemporal dynamics and robustness. Sakhavi \textit{et al.} \cite{sakhavi2018learning} proposed temporal representation approaches to preserve temporal dynamics, while Kwon \textit{et al.} \cite{kwon2020subject} developed subject-independent frameworks leveraging spectral-spatial features for robust calibration-free decoding. Cui \textit{et al.} \cite{cui2023eeg} introduced Interpretable CNNs integrating global average pooling and class activation mapping for consistent pattern identification. For seizure detection, EEGWaveNet \cite{thuwajit2021eegwavenet} leverages dilated temporal convolutions for long-range dependency modeling. While CNNs excel at local pattern extraction, their limited receptive fields constrain global temporal dependency modeling across trial durations.

\subsection{Dual-Stream Parallel Architectures}
Dual-stream networks have emerged as a highly effective architectural paradigm in EEG decoding, explicitly designed to decouple and concurrently process complementary signal aspects via parallel branches, overcoming single-stream model limitations in jointly optimizing local and global dependencies.  Regarding hemispheric asymmetry, Li \textit{et al.} \cite{li2021novel} proposed BiHDM, employing strict dual-stream architecture to process left and right hemisphere signals independently, capturing functional asymmetry. For enhanced spatiotemporal discriminability, TS-SEFFNet \cite{li2022ts} utilizes parallel deep integrated convolutional pathways for concurrent multi-source feature extraction. Addressing subject variability, PCMDA \cite{li2025neuron} leverages parallel contrastive learning streams for simultaneous emotion-specific and domain-invariant representation extraction, while hybrid architectures like DBConformer \cite{wang2025dbconformer} and MVCNet \cite{wang2025mvcnet} combine CNN and Transformer branches to synergize local feature extraction with global dependency modeling. Beyond temporal and spatial characteristics, researchers have expanded this paradigm to spectral and structural domains. Sun \textit{et al.} \cite{sun2022dbgcn} introduced DBGCN, utilizing parallel graph convolution blocks for temporal and spectral representation learning. Similarly, Li \textit{et al.} \cite{Li2025Dual-TSST} proposed dual-branch Transformers where one branch extracts spatiotemporal features from raw signals while another captures spectral dynamics from wavelet-transformed data. Structurally, DS-AGC \cite{Ye2025DS-AGC} employs dual streams for non-structural and structural adaptation via graph contrastive learning, and FBSTCNet \cite{Huang2024FBSTCNet} incorporates parallel branches to jointly decode power-related dynamics and Pearson correlation-based connectivity. However, despite extracting distinct feature sets, these architectures predominantly rely on late fusion, treating streams as isolated information silos until final classification.

\subsection{Feature Fusion Strategies in Multi-Stream Architectures}
Feature fusion strategies in multi-stream EEG architectures have evolved from simple concatenation approaches to sophisticated attention-based mechanisms. Early dual-stream models typically employ basic concatenation or element-wise operations to combine features from parallel branches before final classification \cite{li2021novel, li2022ts}. More advanced approaches utilize weighted fusion schemes, where learnable parameters determine the contribution of each stream \cite{sun2022dbgcn}. Cross-modal attention mechanisms, successfully applied in medical image analysis for fusing CT and MRI modalities \cite{liu2025review}, have been adapted for EEG processing. Within the EEG domain, attention-based fusion has shown promise in integrating heterogeneous representations. Gong \textit{et al.} \cite{Gong2024Bi-Hemi} adopted transformer-based fusion modules to integrate global spatial projection matrices with bi-hemisphere discrepancy streams, modeling dependencies across spatial, spectral, temporal, and hemispheric information. DMSACNN \cite{Liu2025DMSACNN} utilizes local and global feature fusion attention modules to integrate discriminative spatiotemporal patterns at multiple hierarchical levels, while MSVTNet \cite{liu2024msvtnet} leverages Transformer modules to fuse multi-scale spatiotemporal representations from CNN extractors for enhanced motor-imagery decoding.

However, despite these advancements, most existing attention-based frameworks predominantly deploy attention as a late-stage integration module or for static feature weighting. This design effectively treats feature extraction in different streams as isolated processes, neglecting the potential for mutual guidance and refinement at earlier network layers. By deferring interaction to the final classification stage, these models miss critical opportunities to capture the intrinsic, coupled spatiotemporal dynamics of EEG signals throughout the encoding hierarchy. Therefore, developing a progressive, layer-wise interactive architecture that enables continuous cross-stream knowledge exchange, rather than mere final aggregation, represents a critical yet underexplored direction for advancing robust EEG decoding.

\section{Methodology}

\subsection{Architecture Overview}

As illustrated in Figure \ref{fig:ov}, the proposed LI-DSN accepts EEG signal input $\mathbf{X}\in\mathbb{R}^{C\times T}$, where $C$ denotes the number of electrodes and $T$ the temporal samples. The architecture begins with parallel Temporal and Spatial Tokenizers that extract initial embeddings, which then traverse $N$ interactive blocks containing complementary temporal and spatial pathways. Unlike conventional late-fusion designs, LI-DSN employs the Temporal-Spatial Integration Attention (TSIA) mechanism at each layer, enabling spatial structural priors to progressively refine temporal features throughout the encoding hierarchy. Finally, the evolved representations from both streams are synthesized via an Adaptive Fusion module to generate discriminative features for the subsequent multi-layer perceptron (MLP) classifier.

\subsection{Dual-Stream Tokenization}

\subsubsection{Temporal Tokenization}
The temporal stream partitions the input signal into token sequences to capture fine-grained dynamic features. As illustrated in the Temporal Tokenizer module of Figure \ref{fig:ov}, we employ a sequential convolutional pipeline. Let $\mathbf{W}_{pw}$ denote the pointwise convolution weights and $\mathcal{K}_{dw}$ the depth-wise convolution kernels. The token generation process is formulated as:
\begin{equation}
\mathbf{Z}_t = \mathcal{P}_{\downarrow} \Big( \sigma \Big( \mathcal{K}_{dw} \ast \text{BN} \big( \mathbf{W}_{pw} \mathbf{X} \big) \Big) \Big),
\label{eq:temporal_token}
\end{equation}
where $\ast$ represents the convolution operation, $\text{BN}(\cdot)$ is Batch Normalization, and $\sigma(\cdot)$ denotes the GELU activation function. The operator $\mathcal{P}_{\downarrow}$ performs temporal downsampling to reduce resolution, yielding the final embedding $\mathbf{Z}_t\in\mathbb{R}^{P\times D}$, where $P$ denotes the number of temporal patches and $D$ represents the embedding dimension.

\subsubsection{Spatial Tokenization}
\begin{figure*}
    \centering
    \includegraphics[width=1\linewidth]{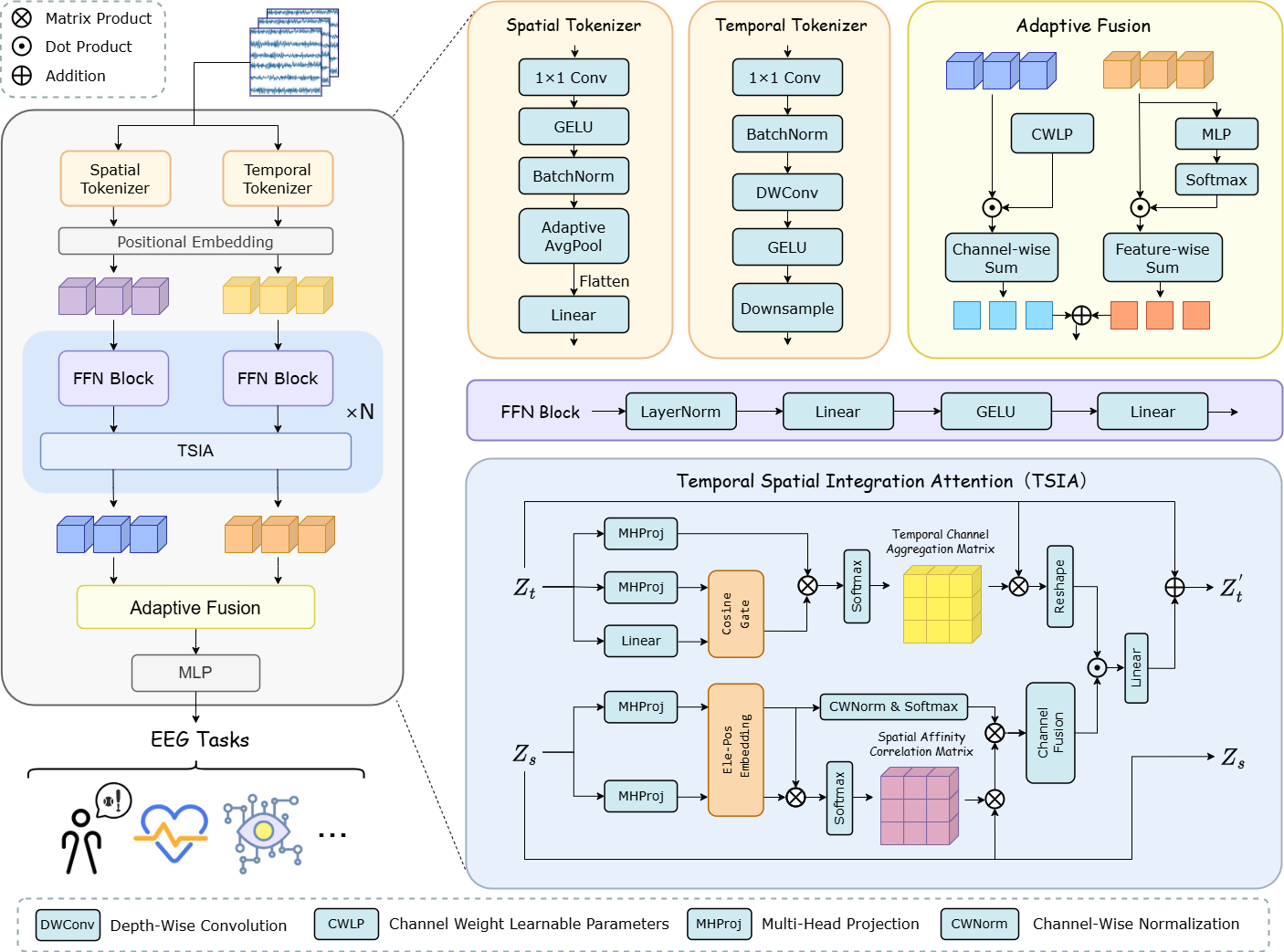}
    \caption{The overall architecture of the proposed LI-DSN.  The raw EEG signals are first processed by parallel Temporal and Spatial Tokenizers to extract initial embeddings.  These embeddings are then fed into $N$ blocks, where each block consists of a Feed-Forward Network (FFN) followed by the Temporal-Spatial Integration Attention (TSIA) module to enable progressive cross-stream interaction.  The refined features are subsequently aggregated by the Adaptive Fusion module and passed through an MLP classification head to generate outputs for various EEG tasks (e.g., MI classification, Emotion Recognition, SSVEP). The detailed architectures of the five key components, namely the Spatial Tokenizer, Temporal Tokenizer, FFN Block, TSIA and Adaptive Fusion, are illustrated in the sub-panels.}
    \label{fig:ov}
\end{figure*}

The spatial stream extracts a compact representation for each electrode. As depicted in the Spatial Tokenizer module of Figure \ref{fig:ov}, each channel's signal $\mathbf{x}_c$ undergoes a dedicated encoding pipeline independent of other channels. We formulate this in two stages. First, we extract the channel-specific feature representation $\mathbf{h}_c$ through a convolution-activation-normalization sequence followed by average pooling:
\begin{equation}
\mathbf{h}_c = \text{AvgPool} \Big( \text{BN} \big( \sigma ( \mathbf{W}_{in} \mathbf{x}_c ) \big) \Big),
\label{eq:spatial_feat}
\end{equation}
where $\mathbf{W}_{in}$ represents the input projection weights. Second, we flatten this representation and map it to the embedding space via a linear transformation:
\begin{equation}
\mathbf{z}_s^{(c)} = \mathbf{W}_{out} \text{vec}(\mathbf{h}_c),
\label{eq:spatial_token}
\end{equation}
where $\mathbf{W}_{out}$ is the linear projection matrix and $\text{vec}(\cdot)$ denotes the flattening operation. This yields the complete spatial embedding $\mathbf{Z}_s\in\mathbb{R}^{C\times D}$, which serves as the initial input for the subsequent interactive blocks.

\subsection{Feed-Forward Network Block}
Before entering the TSIA mechanism, the tokenized features from both streams are first enriched with positional embeddings implemented via learnable parameters to preserve sequence order and spatial structure. These augmented features are then processed through a Feed-Forward Network (FFN) Block to enhance feature representation. As illustrated in Figure \ref{fig:ov}, the FFN Block consists of a Layer Normalization step followed by a residual pathway containing two linear transformations separated by a GELU activation function and a dropout layer. This block refines the features within their respective streams before they undergo cross-stream interaction.

\subsection{Temporal-Spatial Integration Attention (TSIA)}
Following the FFN refinement, the TSIA mechanism facilitates bidirectional guidance between the temporal and spatial domains. As shown in Figure \ref{fig:ov}, it comprises three sequential stages: spatial context extraction via SACM, temporal feature refinement via TCAM, and the final multiplicative integration.

\subsubsection{Spatial Affinity Correlation Matrix (SACM)}
In the spatial stream, the SACM captures the latent correlations between electrodes. We first project the spatial input $\mathbf{Z}_s$ into two feature spaces $\mathbf{Y}_1, \mathbf{Y}_2$ via learnable projection matrices $\mathbf{W}_1, \mathbf{W}_2 \in \mathbb{R}^{D \times d_h}$, where $H$ denotes the number of attention heads and $d_h = D/H$ represents the dimension per head. Simultaneously, we inject learnable electrode position embeddings $\mathbf{E}_{pos} \in \mathbb{R}^{C \times d_h}$ to encode geometric information:
\begin{equation}
\mathbf{Y}_1 = \mathbf{Z}_s \mathbf{W}_1 + \mathbf{E}_{pos}, \quad \mathbf{Y}_2 = \mathbf{Z}_s \mathbf{W}_2 + \mathbf{E}_{pos}.
\label{eq:spatial_proj}
\end{equation}
The SACM is computed to measure the spatial affinity between these representations:
\begin{equation}
\mathbf{A}_{s} = \text{softmax}\left(\frac{\mathbf{Y}_1 \mathbf{Y}_2^\top}{\sqrt{d_h}}\right) \in \mathbb{R}^{H \times C \times C}.
\end{equation}
This affinity matrix is first applied to $\mathbf{Y}_1$ via matrix multiplication to aggregate the spatial context. To emphasize task-relevant channels, we calculate an importance vector $\boldsymbol{\omega} = \text{softmax}(\|\mathbf{Y}_1\|_2)$ based on the L2-norm of the features. The pooled spatial summary $\mathbf{s}_{pool}$ is derived via an importance-weighted summation:
\begin{equation}
\mathbf{s}_{pool} = \sum_{c=1}^C \boldsymbol{\omega}_c (\mathbf{A}_{s} \mathbf{Y}_1)^{(c)}.
\label{eq:sacm_pool}
\end{equation}
Finally, $\mathbf{s}_{pool}$ is broadcast across the temporal dimension $P$ to yield the global spatial context $\mathbf{S}_{exp} \in \mathbb{R}^{P \times D}$.

\subsubsection{Temporal Channel Aggregation Matrix (TCAM)}
In the temporal stream, TCAM models channel-wise dependencies. We first project the temporal input $\mathbf{Z}_t$ into a feature subspace $\mathbf{X}_1 = \mathbf{Z}_t \mathbf{W}_{t1}$, where $\mathbf{W}_{t1} \in \mathbb{R}^{D \times d_h}$ is a learnable projection matrix. Considering that EEG signals are fundamentally composed of neural oscillations with intrinsic rhythmicity, standard monotonic activation functions (e.g., Sigmoid or ReLU) are suboptimal for capturing such periodic dynamics. To explicitly encode these oscillatory properties, we design a periodicity-aware cosine-gated transformation parameterized by $\mathbf{W}_\phi \in \mathbb{R}^{D \times d_h}$:
\begin{equation}
\mathbf{X}_1' = \mathbf{X}_1 \odot \cos(\mathbf{Z}_t \mathbf{W}_\phi).
\end{equation}
The channel-wise correlation matrix is then computed to capture relationships between feature dimensions, utilizing a separate key projection $\mathbf{K}_t = \mathbf{Z}_t \mathbf{W}_{k}^t$:
\begin{equation}
\mathbf{A}_{c} = \text{softmax}\left(\frac{(\mathbf{X}_1')^\top \mathbf{K}_t}{\sqrt{P}}\right) \in \mathbb{R}^{H \times d_h \times d_h}.
\end{equation}
The enhanced temporal features $\widetilde{\mathbf{Z}}_t$ are finally obtained by aggregating the cosine-gated input according to this correlation structure:
\begin{equation}
\widetilde{\mathbf{Z}}_t = \mathbf{X}_1' \mathbf{A}_{c}.
\end{equation}

\subsubsection{Multiplicative Integration}
To synthesize the information from both streams, the refined temporal features $\widetilde{\mathbf{Z}}_t$ and the global spatial context $\mathbf{S}_{exp}$ are fused via element-wise multiplication. This step acts as a soft gating mechanism, where spatial priors highlight relevant temporal patterns. The final output is projected via a linear layer:
\begin{equation}
\mathbf{O}_{\text{TSIA}} = \text{Concat}_{\text{heads}}\big(\widetilde{\mathbf{Z}}_t \odot \mathbf{S}_{\text{exp}}\big)\mathbf{W}_{\text{out}},
\label{eq:tsia_output}
\end{equation}
where $\mathbf{W}_{\text{out}} \in \mathbb{R}^{D \times D}$ denotes the output projection weights.

\subsection{Asymmetric Integration Flow}
At each layer $\ell$, LI-DSN follows an asymmetric update order. First, the spatial stream updates its representations independently through its specific FFN, denoted as $\mathcal{S}^{(\ell)}$. Subsequently, the temporal stream processes its features through a temporal FFN, followed by the interaction with the updated spatial context via the TSIA mechanism. The layer-wise update is formulated as:
\begin{equation}
\begin{aligned}
\mathbf{Z}_s^{(\ell)} &= \mathcal{S}^{(\ell)}\big(\mathbf{Z}_s^{(\ell-1)}\big), \\
\mathbf{H}_t^{(\ell)} &= \text{FFN}\big(\mathbf{Z}_t^{(\ell-1)}\big), \\
\mathbf{Z}_t^{(\ell)} &= \text{TSIA}\big(\mathbf{H}_t^{(\ell)}, \mathbf{Z}_s^{(\ell)}\big).
\end{aligned}
\label{eq:asymmetric_flow}
\end{equation}
This strategy of injecting spatial information into the temporal stream is motivated by the role of the spatial stream as a semantic anchor. Since the temporal dimension contains high-resolution yet redundant dynamics, using compact spatial structural priors to guide temporal feature refinement acts as an effective spatial filter. This unidirectional design enhances the signal-to-noise ratio by filtering out spatially incoherent temporal fluctuations without introducing stochastic noise into the stable spatial structure. As detailed in Section \ref{sec:IM}, empirical results support this architectural choice, indicating that temporal-conditioned integration consistently yields superior performance compared to reverse or symmetric alternatives.

\subsection{Adaptive Fusion}
After $N$ interactive encoding blocks, the network produces high-level temporal features $\mathbf{Z}_t^{(N)} \in \mathbb{R}^{P \times D}$ and spatial features $\mathbf{Z}_s^{(N)} \in \mathbb{R}^{C \times D}$. To effectively synthesize these representations for the final classification, we employ a dual-branch Adaptive Fusion strategy as illustrated in Figure \ref{fig:ov}.

For the spatial stream, we introduce a set of Channel Weight Learnable Parameters (CWLP), denoted as $\mathbf{w}_{sp} \in \mathbb{R}^C$. Unlike standard average pooling, this mechanism allows the model to learn a static importance score for each electrode, explicitly prioritizing task-relevant brain regions (e.g., motor cortex for MI classification tasks). The global spatial representation $\mathbf{z}_{s}$ is obtained via weighted summation:
\begin{equation}
\mathbf{z}_{s} = \sum_{c=1}^{C} w_{sp}^{(c)} \cdot \mathbf{Z}_s^{(N, c)}.
\label{eq:spatial_fusion}
\end{equation}

For the temporal stream, we utilize a dynamic attention-based pooling mechanism to identify discriminative time segments. The temporal features are processed by a MLP consisting of two linear layers with a ReLU activation in between. This MLP maps the $D$-dimensional feature at each time step $p$ to a scalar attention score, which is then normalized via a Softmax function:
\begin{equation}
\boldsymbol{\alpha} = \text{softmax}\big(\mathbf{W}_2 \cdot \text{ReLU}(\mathbf{W}_1 \mathbf{Z}_t^{(N)})\big) \in \mathbb{R}^P,
\label{eq:temporal_att_score}
\end{equation}
where $\mathbf{W}_1$ and $\mathbf{W}_2$ are the weights of the MLP. The final temporal representation $\mathbf{z}_{t}$ is derived by aggregating the temporal tokens according to these learned attention weights:
\begin{equation}
\mathbf{z}_{t} = \sum_{p=1}^{P} \alpha_p \cdot \mathbf{Z}_t^{(N, p)}.
\label{eq:temporal_fusion}
\end{equation}

Finally, the aggregated spatial and temporal vectors are concatenated to form the unified representation $\mathbf{u} = [\mathbf{z}_{t}; \mathbf{z}_{s}] \in \mathbb{R}^{2D}$. This vector is fed into the final MLP classifier to obtain the predicted class probabilities $\hat{y}$. The model is trained end-to-end using standard cross-entropy loss:
\begin{equation}
\mathcal{L}_{\text{CE}} = -\frac{1}{N}\sum_{i=1}^{N} \log\big(\hat{y}_{i},{y_i}\big),
\label{eq:loss}
\end{equation}
where $N$ is the batch size and $\hat{y}_{y_i}$ corresponds to the predicted probability of the ground truth class $y_i$.

\begin{algorithm}[t]
\caption{Training Procedure of LI-DSN}
\label{alg:training}
\begin{algorithmic}[1]
\REQUIRE Training set $\mathcal{D}_{\text{train}}$, batch size $B$, max epochs $E$, class weights $\mathbf{w}$
\ENSURE Optimized parameters $\boldsymbol{\theta}$
\STATE Initialize model components and Adam optimizer ($\beta=(0.9, 0.95)$, weight decay $=0$)
\FOR{$\text{epoch} = 1$ to $E$}
    \FOR{each batch $\{(\mathbf{X}_i, y_i)\}_{i=1}^{B}$ in $\mathcal{D}_{\text{train}}$}
        \STATE \textbf{// Dual-stream Tokenization}
        \STATE $\mathbf{Z}_t^{(0)} \leftarrow \text{TemporalTokenizer}(\mathbf{X})$ via Eq.~\eqref{eq:temporal_token}
        \STATE $\mathbf{Z}_s^{(0)} \leftarrow \text{SpatialTokenizer}(\mathbf{X})$ via Eq.~\eqref{eq:spatial_token}
        
        \STATE \textbf{// Layer-wise Interactive Encoding}
        \FOR{$\ell = 1$ to $N$}
            \STATE Update $\mathbf{Z}_s^{(\ell)}$ and $\mathbf{Z}_t^{(\ell)}$ via Asymmetric Flow (Eq.~\eqref{eq:asymmetric_flow})
        \ENDFOR
        
        \STATE \textbf{// Fusion and Optimization}
        \STATE $\mathbf{u} \leftarrow \text{AdaptiveFusion}(\mathbf{Z}_t^{(N)}, \mathbf{Z}_s^{(N)})$ via Eq.~\eqref{eq:spatial_fusion}~\eqref{eq:temporal_fusion}
        \STATE $\hat{\mathbf{y}} \leftarrow \text{MLP}_{\text{cls}}(\mathbf{u})$
        \STATE $\mathcal{L} \leftarrow -\frac{1}{B}\sum_{i=1}^{B} w_{y_i} \log(\hat{y}_{i,y_i})$ via Eq.~\eqref{eq:loss}
        \STATE Update $\boldsymbol{\theta}$ via backpropagation: $\boldsymbol{\theta} \leftarrow \boldsymbol{\theta} - \eta \nabla_{\boldsymbol{\theta}} \mathcal{L}$
    \ENDFOR
    \STATE Evaluate accuracy on $\mathcal{D}_{\text{test}}$
\ENDFOR
\RETURN $\boldsymbol{\theta}$
\end{algorithmic}
\end{algorithm}

\section{Experiment}
\subsection{Datasets}
We evaluated the performance of LI-DSN across three distinct EEG decoding tasks: MI classification, emotion recognition, and SSVEP. The characteristics of the datasets utilized are detailed in Table~\ref{tab:datasets}.
\subsubsection{MI classification Datasets}
    \begin{itemize}
        \item BNCI2014001~\cite{bnci2014001}: Nine subjects performed left and right hand MI classification tasks while 22 EEG channels were recorded at 250 Hz over 4~s trials. Only the calibration session was used, yielding 144 trials per subject and a total of 1,296 trials across all subjects.
        \item BNCI2014002~\cite{bnci2014002}: Fourteen subjects performed right hand and feet MI classification tasks while 15 EEG channels recorded at 512 Hz. Each subject contributed 100 trials from the first five runs, yielding 1,400 trials in total.
        \item BNCI2014004~\cite{bnci2014004}: Nine subjects performed left and right hand MI classification tasks, recorded from 3 EEG channels at 250 Hz. We use only the first session, yielding 120 trials per subject and 1,080 trials in total.
        \item Zhou2016~\cite{zhou2016}: Four subjects performed left and right hand MI classification tasks, recorded from 14 EEG channels at 250 Hz. We use only the first session, yielding 409 usable 5~s trials in total.
    \end{itemize}

\subsubsection{Emotion Recognition Datasets}
    \begin{itemize}
        \item SEED~\cite{seed}: Fifteen subjects watched 15 film clips across three sessions to elicit positive, neutral, and negative emotions, while 62 EEG channels were recorded at 1,000 Hz (down-sampled to 200~Hz). After discarding the neutral class and segmenting the data into 2~s windows, we obtained 7,920 labeled trials for binary classification.
        \item FACED~\cite{faced}: One hundred and twenty-three subjects viewed 28 video clips to elicit 9 emotions, while 32 EEG channels were recorded at 250 or 1000~Hz (down-sampled to 250~Hz). The resulting valence scores were binarized into high and low classes using a threshold of 3.0, and the EEG signals were segmented into 4~s windows, yielding 10,332 labeled trials in total.
    \end{itemize}

\subsubsection{SSVEP Datasets}
    \begin{itemize}
        \item Nakanishi2015~\cite{nakanishi2015}: Nine subjects focused on 12 flickering targets for 4~s while eight occipital electrodes were recorded at 2048~Hz (down-sampled to 256~Hz). Each subject contributes 180 trials across all stimulus codes, yielding 1,620 trials in total.
        \item Kalunga2016~\cite{kalunga2016}: Twelve subjects performed a 4-class SSVEP task (13/17/21~Hz plus resting). EEG was acquired with eight channels at 256~Hz. Every session contains 32 trials (5~s stimulation). Across 2–5 sessions per subject, a total of 945 labeled trials were collected.
     \end{itemize}
     \begin{table*}[!t]
\caption{MI classification accuracies (\%) of different methods on the BNCI2014001, BNCI2014002, BNCI2014004, and Zhou2016 datasets under CO, CV, and LOSO scenarios. (\textbf{Bold}: best; \underline{Underline}: second best)}
\centering
\renewcommand{\arraystretch}{1.5}
\setlength{\tabcolsep}{1.1pt}
\begin{tabular}{c|c|cccccccccc}
\hline
\hline
Task & Dataset & SCNN & DCNN & EEGNet & FBCNet & CTNet & MSVTNet & IFNet & Conformer & DBConformer & \textbf{LI-DSN} \\
\hline
\multirow{4}{*}{\centering CO} & BNCI2014001  & 73.57$\pm$2.37 & 59.29$\pm$1.64 & 69.05$\pm$1.02 & 68.97$\pm$1.28 & 73.49$\pm$2.12 & 74.60$\pm$3.03 & 77.94$\pm$0.87 & 78.57$\pm$0.72 & \underline{80.16$\pm$1.83} & \textbf{80.95$\pm$1.09} \\
& BNCI2014002 & 79.07$\pm$2.01 & 64.07$\pm$2.68 & 66.07$\pm$2.76 & 69.50$\pm$1.02 & 71.00$\pm$1.78 & 78.93$\pm$1.42 & 78.29$\pm$1.73 & 76.21$\pm$1.52 &  \underline{79.14$\pm$0.16} & \textbf{81.57$\pm$1.33} \\
& BNCI2014004 & 68.15$\pm$1.42 & 62.41$\pm$1.58 & 68.43$\pm$2.21 & 65.46$\pm$1.56 & 71.57$\pm$1.32 & 71.30$\pm$1.50 & 73.43$\pm$1.71 & 72.04$\pm$1.61 & \underline{75.09$\pm$1.54} & \textbf{75.19$\pm$0.23} \\
& Zhou2016 & 75.03$\pm$6.18 & 78.03$\pm$2.41 & 80.13$\pm$3.35 & 63.33$\pm$2.31 & 76.81$\pm$5.14 & 70.32$\pm$1.42 & 81.70$\pm$2.14 & 73.87$\pm$4.45 & \underline{82.49$\pm$2.42} & \textbf{84.65$\pm$1.87} \\

\hline
\multirow{4}{*}{\centering CV} & BNCI2014001 & 78.22$\pm$0.21 & 61.59$\pm$1.12 & 71.86$\pm$1.03 & 74.74$\pm$0.72 & 75.97$\pm$0.77 & 81.28$\pm$0.72 & 82.62$\pm$0.92 & 83.62$\pm$1.01 & \underline{83.66$\pm$0.42} & \textbf{83.79$\pm$0.69} \\
& BNCI2014002 & 81.24$\pm$0.52 & 67.17$\pm$0.32 & 68.90$\pm$1.37 & 74.53$\pm$0.72 & 75.50$\pm$0.93 & 80.91$\pm$0.52 & 80.21$\pm$0.52 & 79.30$\pm$0.31 & \underline{81.36$\pm$0.33} & \textbf{82.57$\pm$0.44} \\
& BNCI2014004 & 68.13$\pm$0.61 & 63.51$\pm$0.42 & 68.91$\pm$0.88 & 67.17$\pm$0.31 & 70.96$\pm$1.11 & 73.59$\pm$0.64 & 72.02$\pm$0.53 & 73.15$\pm$0.71 & \underline{74.17$\pm$0.62} & \textbf{75.11$\pm$0.18} \\
& Zhou2016 & 82.90$\pm$0.91 & 79.68$\pm$1.11 & 85.30$\pm$2.12 & 80.01$\pm$0.52 & 86.38$\pm$1.48 & 88.06$\pm$1.21 & 87.78$\pm$1.01 & 85.67$\pm$1.52 & \textbf{91.54$\pm$0.17} & \underline{91.48$\pm$1.38} \\

\hline
\multirow{4}{*}{\centering LOSO} & BNCI2014001 & 72.22$\pm$1.02 & 73.21$\pm$1.81 & 73.64$\pm$1.11 & 72.56$\pm$1.02 & 73.40$\pm$1.22 & 77.10$\pm$1.33 & 74.52$\pm$0.67 & 73.07$\pm$2.01 & \underline{77.67$\pm$0.55} & \textbf{77.86$\pm$1.23} \\
& BNCI2014002 & 70.57$\pm$1.41 & 74.34$\pm$0.81 & 72.86$\pm$0.41 & 71.31$\pm$0.77 & 74.14$\pm$0.81 & 74.61$\pm$1.12 & 73.90$\pm$0.71 & 72.84$\pm$1.44 & \underline{77.17$\pm$0.84} & \textbf{77.25$\pm$0.13} \\
& BNCI2014004 & 62.19$\pm$0.80 & 62.56$\pm$0.62 & 67.78$\pm$0.90 & 67.09$\pm$0.74 & 65.28$\pm$1.09 & 67.41$\pm$0.83 & 67.69$\pm$0.39 & 64.22$\pm$1.14 & \underline{69.85$\pm$0.52} & \textbf{71.44$\pm$0.27} \\
& Zhou2016 & 82.10$\pm$0.71 & 83.84$\pm$1.21 & 83.22$\pm$1.80 & 82.07$\pm$0.93 & 83.88$\pm$1.01 & 84.95$\pm$1.44 & \underline{86.21$\pm$0.97} & 82.43$\pm$1.46 & 85.37$\pm$0.78 & \textbf{86.53$\pm$1.24} \\

\hline
\hline
\end{tabular}
\label{tab:mi}
\end{table*}

\subsection{Baselines}
We evaluate the performance of our proposed LI-DSN by comparing it with the following 13 baseline methods:

\begin{itemize}
\item SCNN (Shallow Convolutional Neural Network) \cite{schirrmeister2017deep}: A shallow CNN architecture for EEG decoding and inspired by the Filter Bank Common Spatial Pattern (FBCSP) paradigm. It employs temporal convolution for frequency-specific filtering and spatial convolution across EEG channels to extract discriminative spatiotemporal patterns, enabling effective decoding of local EEG dynamics.
\item DCNN (Deep Convolutional Neural Network) \cite{schirrmeister2017deep}: A deeper version of SCNN with four convolutional blocks and max pooling layers for feature extraction. Its increased depth enables capturing more complex features and intricate patterns in EEG signals.
\item GCBNet (Graph Convolutional Broad Network) \cite{zhang2019gcb}: A hybrid framework integrating Graph Convolutional Networks (GCN) to model the topological structure of EEG channels, with stacked standard convolutional layers. It employs a broad learning strategy by concatenating hierarchical features across layers, efficiently capturing both local graph interactions and high-level abstract patterns.
\item EEGNet (Compact Convolutional Neural Network) \cite{lawhern2018eegnet}: A compact CNN architecture for generalized EEG classification. It utilizes depthwise separable convolutions to efficiently extract frequency-specific temporal features and spatial filters, reducing trainable parameters while enhancing spatiotemporal representation capabilities.
\item TSception (Multi-scale Convolutional Neural Network) \cite{ding2022tsception}: A multi-scale CNN that models EEG’s temporal dynamics and hemispheric asymmetry using dynamic temporal kernels and asymmetric spatial convolutions, providing a strong baseline for robust spatiotemporal emotion recognition.
\item FBCNet (Filter Bank Convolutional Neural Network) \cite{mane2021fbcnet}: A multi-view convolutional network inspired by the FBCSP pipeline. It decomposes EEG signals into multiple frequency bands and employs spatial convolutional layers followed by log-variance layers to extract discriminative spectral-spatial features, avoiding the computational complexity of standard deep ConvNets.
\item CTNet (Convolutional Transformer Network) \cite{zhao2024ctnet}: A hybrid framework that synergizes the local inductive bias of CNNs with the global context modeling capabilities of Transformers. A convolutional front-end extracts fine-grained local patterns, which are then processed by a Transformer encoder to capture long-range temporal dependencies and global interactions.
\item MSVTNet (Multi-Scale Vision Transformer Network) \cite{liu2024msvtnet}: A hierarchical framework that integrates multi-scale CNNs with Transformers to capture cross-frequency coupling and global correlations. It employs auxiliary branch loss for intermediate supervision, ensuring robust capture of spatio-temporal dynamics and cross-scale interactions.
\item IFNet (Interactive Frequency Convolutional Neural Network) \cite{wang2023ifnet}: A multi-branch architecture that decomposes EEG into specific frequency bands processed by 1D convolutions. It employs an attention-based fusion mechanism to facilitate cross-frequency interaction, effectively capturing complementary spectral patterns that are often overlooked by single-stream models.
\item EEG-Conformer (Compact Convolutional Transformer) \cite{song2023eeg}: A compact, unified framework composed of a convolutional module followed by a Transformer encoder, synergistically combining the local spatiotemporal feature extraction capabilities of CNNs with the global long-range dependency modeling of attention mechanisms.
\item EEG-Deformer (Dense Convolutional Transformer) \cite{ding2024eeg}: A hierarchical framework designed to mitigate the loss of local details in standard Transformers. It integrates a Hierarchical Coarse-to-Fine Transformer with a parallel Fine-grained Temporal Learning branch, enabling the simultaneous capture of macroscopic temporal trends and microscopic signal dynamics.
\item DBConformer (Dual-Branch Convolutional Transformer) \cite{wang2025dbconformer}: A parallel dual-branch architecture designed to capture global EEG representations efficiently. It integrates a temporal Conformer with a spatial Conformer and a lightweight channel attention module, enabling the simultaneous modeling of long-range temporal dependencies and intricate inter-channel interactions with high interpretability and minimal parameter overhead.
\item EmT (Emotion Transformer) \cite{ding2025emt}: A robust framework tailored for cross-subject emotion recognition. It models EEG signals as temporal graphs and utilizes a residual multiview pyramid GCN to learn topological representations, while a Temporal Contextual Transformer captures the evolution of these emotional states over time.
\end{itemize}

\subsection{Experiment Settings}

To comprehensively evaluate the proposed LI-DSN model, we conducted experiments across three BCI paradigms: MI classification, emotion recognition, and SSVEP. For MI classification tasks, we employed four public datasets (BNCI2014001, BNCI2014002, BNCI2014004, and Zhou2016) and evaluated our model under three scenarios: Chronological Order (CO), Cross-Validation (CV), and Leave-One-Subject-Out (LOSO). In CO, trials were split chronologically, with the first 80\% for training and the last 20\% for testing, simulating a realistic time-ordered deployment. CV was performed using 5-fold cross-validation by splitting each subject’s data into five contiguous time-ordered segments. LOSO tested cross-subject generalization by training on all subjects except one. For emotion recognition and SSVEP, we used the LOSO protocol to evaluate cross-subject generalization, while for the FACED dataset, a 10-fold cross-subject validation was employed. For MI classification and SSVEP datasets, each experiment was conducted five times using different random seeds, and the mean classification accuracy along with standard deviation across subjects was reported. For emotion recognition, results were obtained without repeated random initialization.

\begin{table}[!htbp]
    \caption{Emotion recognition results of different methods on the SEED and FACED datasets. (\textbf{Bold}: best; \underline{Underline}: second best)}
    \renewcommand{\arraystretch}{1.4}
    \centering
    \footnotesize
    \setlength{\tabcolsep}{1pt}
    \scalebox{1}{
    \begin{tabular}{lcccc}
        \hline
        \hline
        \multirow{2}{*}{Method} & \multicolumn{2}{c}{SEED} & \multicolumn{2}{c}{FACED} \\
        \cline{2-5}
        & ACC(±std)↑ & F1(±std)↑ & ACC(±std)↑ & F1(±std)↑ \\
        \hline
        GCBNet & $68.42\pm17.21$ & $52.15\pm35.68$ & $57.38\pm5.23$ & $69.12\pm5.36$ \\
        EEGNet & $69.55\pm11.28$ & $65.84\pm22.08$ & $62.60\pm4.35$ & $76.45\pm3.31$ \\
        TSception & $66.23\pm18.12$ & $62.05\pm28.26$ & $61.92\pm8.83$ & $70.19\pm3.58$ \\
        CTNet & $68.71\pm11.62$ & $63.75\pm25.41$ & $62.08\pm4.20$ & $76.10\pm3.17$ \\
        MSVTNet & $64.36\pm8.21$ & $59.86\pm21.78$ & \underline{$62.91\pm6.17$} & \underline{$76.79\pm4.84$} \\
        Conformer & $61.58\pm12.70$ & $52.98\pm22.04$ & $59.06\pm3.53$ & $72.36\pm3.58$ \\
        Deformer & $75.95\pm10.68$ & $74.72\pm12.39$ & $57.64\pm7.08$ & $69.15\pm7.01$ \\
        EmT & \underline{$80.42\pm11.56$} & $\mathbf{82.14\pm9.38}$ & $60.81\pm6.58$ & $74.55\pm5.85$ \\
        \hline
        LI-DSN & $\mathbf{82.37\pm11.05}$ & \underline{$81.34\pm18.65$} & $\mathbf{62.99\pm4.41}$ & $\mathbf{77.04\pm4.02}$ \\
        \hline
        \hline
    \end{tabular}}
    \label{seedfaced}
\end{table}

\subsection{Implementation Details}
The proposed LI-DSN model was implemented using PyTorch and trained on NVIDIA GeForce RTX 4090 GPUs with multi-GPU support. The model architecture used an embedding dimension of 40, spatial dimension of 16, Transformer encoders with 3 layers for temporal and channel branches, and dropout rate of 0.25. Training hyperparameters were consistent across all scenarios: learning rate of 0.001, batch size of 32, and maximum 100 epochs with early stopping mechanism to prevent overfitting.  We employed the Adam optimizer with momentum parameters (\(\beta_{1}\)=0.9, \(\beta_{2}\)=0.95), epsilon value of 1e-8 for numerical stability, and no weight decay. 

For data preprocessing, we applied dataset-specific protocols aligned with our implementation. MI classification datasets utilized MOABB-provided signals without additional band-pass filtering or notch filters, following standard session and class trimming. Euclidean Alignment (EA) was applied by default to mitigate inter-subject covariance variability. 
For emotion recognition datasets, we used the officially preprocessed data. A two-stage temporal segmentation was applied to capture both coarse- and fine-grained dynamics. First, 20~s sliding windows with 0.8 overlap partitioned the EEG signals into segments. Each segment was further divided into sub-segments using dataset-specific windows (4~s for FACED, 2~s for SEED) with 0.75 overlap. For each sub-segment, relative power spectral density (rPSD) features were extracted across seven frequency bands ($\delta$: 1–3~Hz, $\theta$: 4–8~Hz, $\alpha$: 8–12~Hz, $\beta$: 12–16~Hz, low-$\gamma$: 16–20~Hz, mid-$\gamma$: 20–28~Hz, high-$\gamma$: 30–45~Hz) and normalized per channel using $z$-score. Finally, the sub-segment and frequency dimensions were concatenated for each channel, yielding a compact feature vector encoding local temporal and spectral information.
For SSVEP datasets, signals were resampled to 256 Hz via MOABB with time windowing and without additional filtering. EA was disabled for Nakanishi2015 while retained for Kalunga2016.

\subsection{Evaluation Metrics}

To assess the proposed LI-DSN model across different BCI paradigms, we employed task-specific evaluation metrics. For MI classification tasks, we used classification accuracy (ACC):
\begin{equation}
\text{ACC} = \frac{N_c}{N_t} \times 100\%
\end{equation}
where $N_c$ is the number of correct predictions and $N_t$ is the total number of predictions.

For emotion recognition and SSVEP tasks, we reported ACC and F1-score, defined as:
\begin{equation}
\text{F1} = 2 \times \frac{P \times R}{P + R}
\end{equation}
where $P = \frac{\text{TP}}{\text{TP} + \text{FP}}$ and $R = \frac{\text{TP}}{\text{TP} + \text{FN}}$ are precision and recall, with TP, FP, FN denoting true positives, false positives, and false negatives. All metrics were computed per-subject and averaged across subjects with standard deviations reported.

\begin{table}[!htbp]
    \caption{SSVEP results of different methods on the Nakanishi2015 and Kalunga2016 datasets. (\textbf{Bold}: best; \underline{Underline}: second best)}
    \renewcommand{\arraystretch}{1.3}
    \centering
    \footnotesize
    \setlength{\tabcolsep}{1pt}
    \scalebox{1}{
    \begin{tabular}{lcccc}
        \hline
        \hline
        \multirow{2}{*}{Method} & \multicolumn{2}{c}{Nakanishi2015} & \multicolumn{2}{c}{Kalunga2016} \\
        \cline{2-5}
        & ACC(±std)↑ & F1(±std)↑& ACC(±std)↑ & F1(±std)↑ \\
        \hline
        EEGNet & $96.09\pm0.22$ & $95.94\pm0.22$ & $69.73\pm0.68$ & $67.87\pm0.94$ \\
        CTNet & $96.19\pm0.11$ & $96.06\pm0.14$ & $69.18\pm0.45$ & $67.05\pm0.30$ \\
        MSVTNet & $90.46\pm0.27$ & $90.37\pm0.32$ & $70.84\pm0.49$ & $68.35\pm0.54$ \\
        Conformer & $94.16\pm0.34$ & $94.16\pm0.34$ & $65.17\pm0.42$ & $63.10\pm0.52$ \\
        Deformer & $\underline{96.26\pm0.33}$ & $\underline{96.19\pm0.34}$ & $70.01\pm0.35$ & $\underline{68.97\pm0.51}$ \\
        DBConformer & $94.12\pm0.24$ & $93.98\pm0.23$ & $\underline{72.00\pm0.85}$ & $\mathbf{69.97\pm0.98}$ \\
        \hline
        LI-DSN & $\mathbf{97.04\pm0.14}$ & $\mathbf{96.89\pm0.12}$ & $\mathbf{73.10\pm0.49}$ & $68.58\pm0.68$ \\
        \hline
        \hline
    \end{tabular}}
    \label{tab:ssvep}
\end{table}

\section{Results and Discussion}
\subsection{Performance Comparison}

\subsubsection{Motor Imagery Classification}

The experimental results for MI classification are summarized in Table \ref{tab:mi}. We evaluated the methods across Chronological Order (CO), Cross-Validation (CV), and Leave-One-Subject-Out (LOSO) settings on the BNCI2014001, BNCI2014002, BNCI2014004, and Zhou2016 datasets. As shown in Table \ref{tab:mi}, LI-DSN consistently achieves superior performance. On BNCI2014001, it achieves the highest accuracy of 80.95$\pm$1.09\% (CO), 83.79$\pm$0.69\% (CV), and 77.86$\pm$1.23\% (LOSO). For BNCI2014002, it reaches 81.57$\pm$1.33\% (CO) and 82.57$\pm$0.44\% (CV) and 77.25$\pm$0.13\% (LOSO). On BNCI2014004, which has limited channel density, LI-DSN demonstrates robustness with 75.19$\pm$0.23\% (CO) 75.11$\pm$0.18\% and 71.44$\pm$0.27\% (LOSO). Furthermore, on Zhou2016, it achieves $84.65\% \pm 1.87\%$ (CO) and $86.53\% \pm 1.24\%$ (LOSO), setting new SOTA performance while maintaining competitiveness in CV with $91.48\% \pm 1.38\%$.

These results indicate the effectiveness of the proposed layer-wise interactive mechanism. Unlike other dual-stream networks that typically process streams independently or fuse them only at the end, LI-DSN explicitly models the exchange between temporal dynamics and spatial arrangement at multiple levels. This allows the network to refine features progressively. Consequently, the model delivers robust performance even when generalizing to unseen subjects or operating with limited channel information.

\begin{table}[!htbp]
    \caption{The ablation study of the proposed module and strategy on three MI classification datasets.}
    \renewcommand{\arraystretch}{1.4}
    \centering
    \footnotesize
    \setlength{\tabcolsep}{3.5pt}
    \scalebox{1.05}{
    \begin{tabular}{l|ccc}
    \hline
    \hline
    Model Variant & 2014001 & 2014002 & Zhou2016 \\
    \hline
    w/o Positional Embedding & 74.79±1.18 & 74.67±0.14 & 84.87±1.19 \\
    w/o Interaction (Late Fusion) & 73.25±1.32 & 72.05±0.15 & 81.51±1.38 \\
    w/o TSIA (Simple Concat) & 72.37±1.72 & 70.56±0.23 & 80.31±1.65 \\
    w/o Cosine Gate & 75.22±1.15 & 75.29±0.08 & 84.01±1.21 \\
    w/o Ele-Pos Embedding & 75.65±1.20 & 75.34±0.12 & 83.92±1.26 \\
    w/o Adaptive Fusion & 73.69±1.45 & 73.90±0.14 & 82.14±1.41 \\
    \textbf{LI-DSN (Full Model)} & \textbf{77.86±1.23} & \textbf{77.25±0.13} & \textbf{86.53±1.24} \\
    \hline
    \hline
  \end{tabular}}
    \label{tab:ablation}
\end{table}

\subsubsection{Emotion Recognition}

Table~\ref{seedfaced} presents the emotion recognition results. On SEED, LI-DSN achieves the highest ACC of $82.37\pm11.05\%$ and the second-best F1 of $81.34\pm18.65\%$, surpassing the runner-up EmT ($80.42\pm11.56\%$ ACC, $82.14\pm9.38\%$ F1). Results on FACED further validate our model, where it attains top performance with an ACC of $62.99\pm4.41\%$ and F1 of $77.04\pm4.02\%$. Crucially, this represents a substantial improvement over the strongest baseline, MSVTNet ($62.91\pm6.17\%$ ACC, $76.79\pm4.84\%$ F1). This notable margin is attributed to LI-DSN's layer-wise interactive paradigm, which explicitly decouples temporal dynamics and spatial arrangement while facilitating progressive cross-stream exchange. Unlike standard late-fusion architectures, this design effectively captures complex affective dependencies, ensuring high accuracy despite inherent inter-subject variability.

\subsubsection{Steady-State Visually Evoked Potentials}
The robustness of LI-DSN extends to frequency-modulated SSVEP tasks, as evidenced by the results on the Nakanishi2015 and Kalunga2016 datasets (Table \ref{tab:ssvep}). Our model establishes a new SOTA with ACC of $97.04\pm0.14\%$ and $73.10\pm0.49\%$, respectively. It consistently  marking a clear improvement over competitive baselines like Deformer and DBConformer. This performance gain is primarily driven by the TSIA mechanism, which dynamically aligns the frequency oscillations with their corresponding occipital spatial patterns, thereby enhancing the detection of visual evoked potentials amidst background noise.

\begin{figure}
    \centering
    \setlength{\abovecaptionskip}{1pt}
    \includegraphics[width=0.8\linewidth]{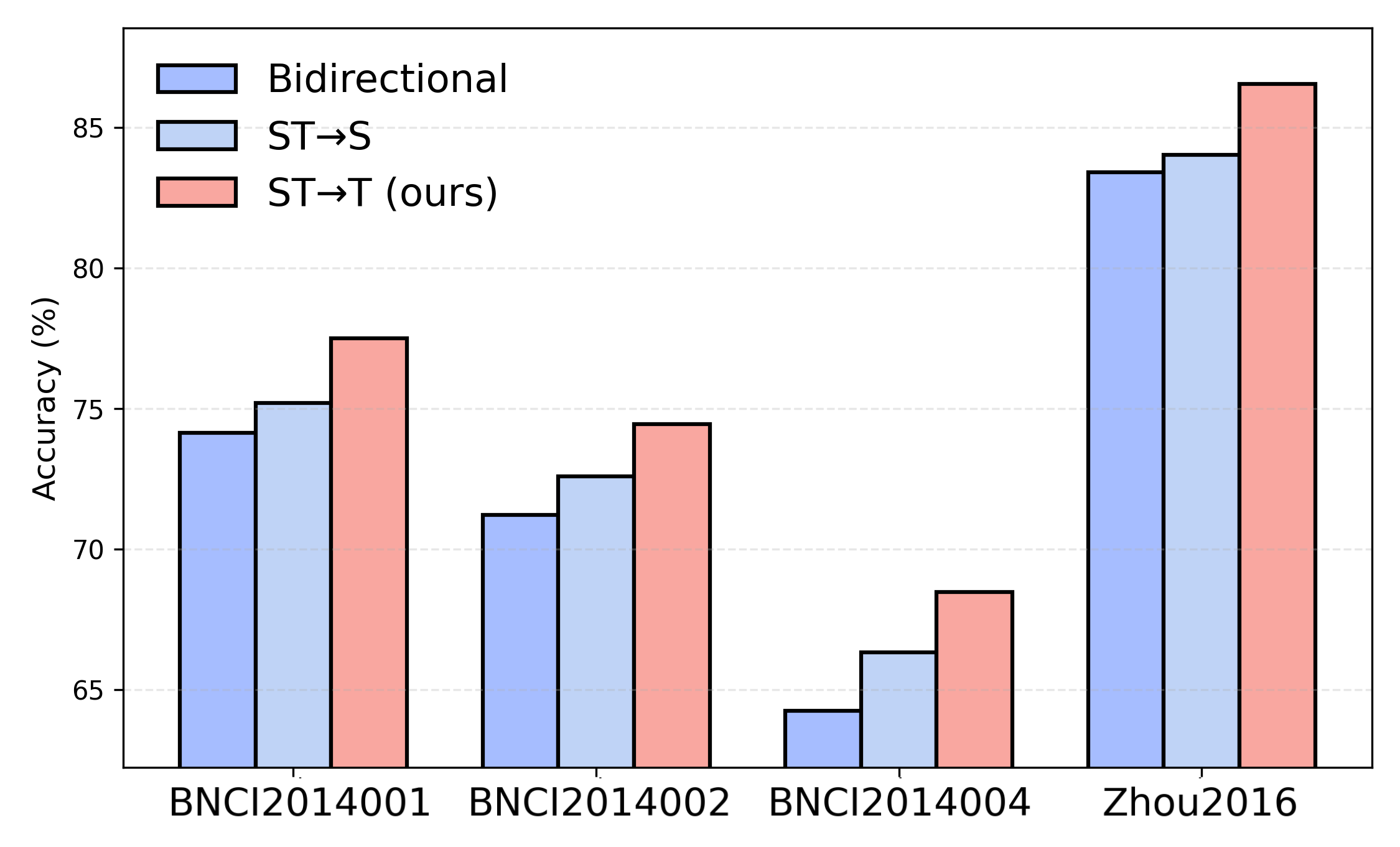}
    \caption{Comparison of integration method under the LOSO protocol across four MI classification datasets.}
    \label{fig:MI vis}
\end{figure}

\subsection{Ablation Study}
To rigorously validate the effectiveness of the core components in LI-DSN, we conducted comprehensive ablation studies on three representative datasets: BNCI2014001, BNCI2014002, and Zhou2016. We specifically investigated the impact of the layer-wise interactive mechanism, the directionality of information flow, and the adaptive fusion strategy. The comparative results are summarized in Table~\ref{tab:ablation}.

\subsubsection{Impact of Positional Embedding}
We investigated the necessity of Positional Embeddings (PE) for encoding sequence order. As shown in Table~\ref{tab:ablation}, removing PE results in a noticeable performance decline, demonstrating that explicit sequence information is vital for the model to interpret dynamic EEG patterns correctly. Without PE, the self-attention mechanism treats the input as a bag of features, losing critical temporal context. We also observed that using learnable positional embeddings provided a slight edge over fixed sinusoidal ones, likely due to their adaptability to the specific temporal distortions present in the dataset.

\subsubsection{Impact of Layer-wise Interaction}

We rigorously evaluated the TSIA module through four variants covering both interaction strategies and internal components: w/o Interaction (Late Fusion),'' w/o TSIA (Simple Concat),'' w/o Cosine Gate,'' and w/o Ele-Pos Embedding,'' where the latter denotes Electrode Position Embedding. As detailed in Table~\ref{tab:ablation}, discarding layer-wise interaction or replacing TSIA with simple concatenation leads to significant performance drops (e.g., $>3\%$ on BNCI2014001), validating the necessity of progressive coupling. Crucially, ablating specific internal components also degrades performance: removing the Cosine Gate or Electrode Position Embedding reduces accuracy to 75.22\% and 75.65\% on BNCI2014001, respectively. These results confirm that beyond the interaction paradigm itself, the specific designs of cosine-gated oscillation modeling and explicit spatial priors are essential for capturing subtle spatiotemporal dynamics.

\begin{figure}
    \centering
    \setlength{\abovecaptionskip}{0.1pt}
    \includegraphics[width=1\linewidth]{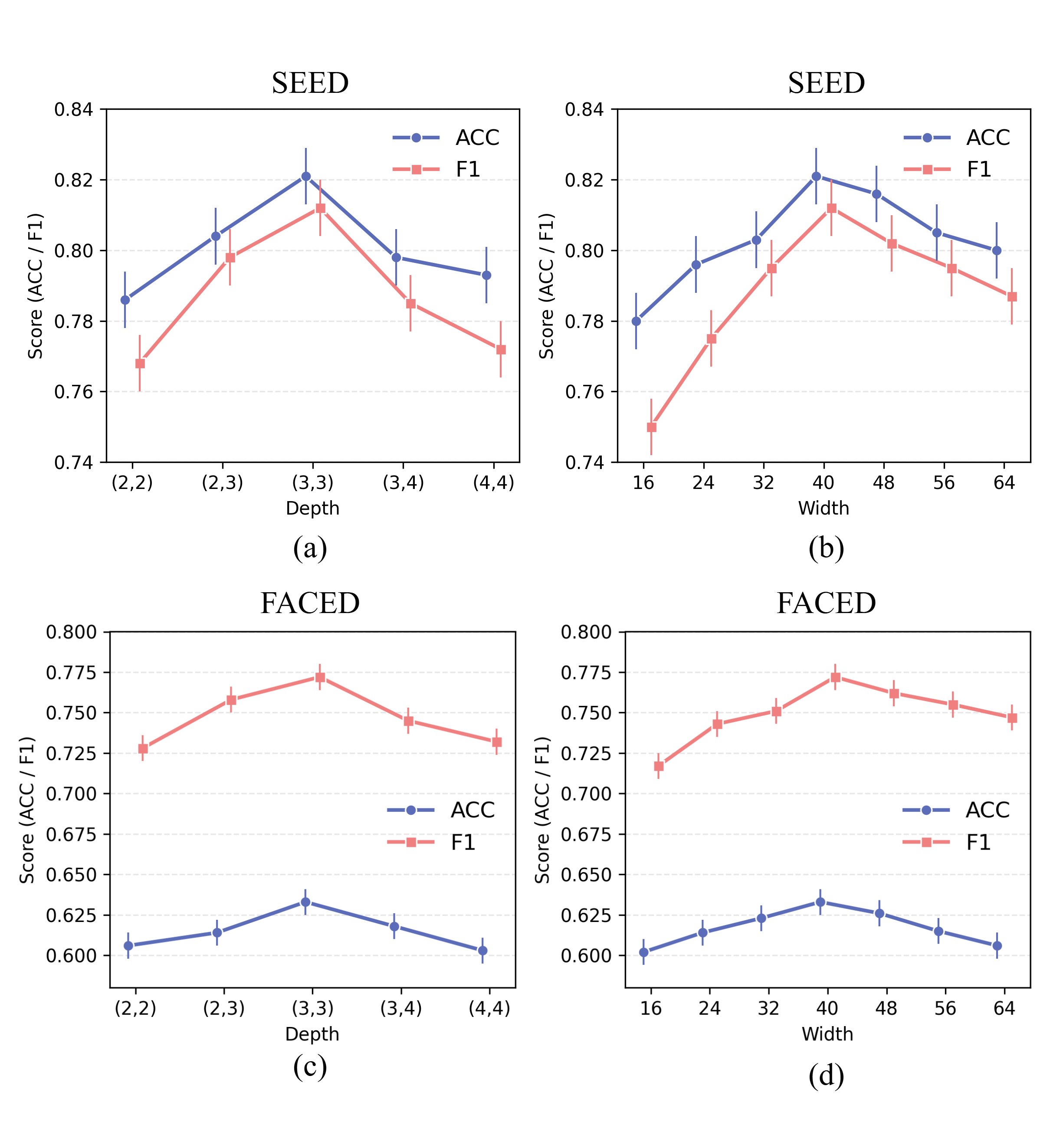}
    \caption{Impact of architectural hyperparameters on performance under the LOSO protocol.   (a) and (c) illustrate the effect of block depth $(N_s, N_t)$ on the SEED and FACED datasets respectively.   (b) and (d) display the effect of feature embedding width on the SEED and FACED datasets.}
    \label{fig:depth_width}
\end{figure}

\subsubsection{Impact of Adaptive Fusion}
Finally, we evaluated the contribution of the Adaptive Fusion Head by replacing it with a standard concatenation of global average pooled features (denoted as ``w/o Adaptive Fusion''). The consistent performance drop in this variant highlights the importance of the learnable channel weights and temporal attention pooling. These components allow the model to dynamically suppress task-irrelevant background noise and focus on the most discriminative time windows and brain regions, which is particularly beneficial for datasets with high inter-subject variability like BNCI2014002.

\subsection{Effect of Integration method}\label{sec:IM}
To further justify our asymmetric spatiotemporal-to-temporal (ST$\rightarrow$T) integration, we compared it with two alternatives: symmetric bidirectional interaction and spatiotemporal-to-spatial (ST$\rightarrow$S), under the LOSO protocol on four MI classification datasets. As illustrated in Figure \ref{fig:MI vis}, in the bidirectional variant, both streams mutually update each layer, while in ST$\rightarrow$S, the temporal stream refines spatial features instead. Across the datasets, bidirectional interaction consistently under-performs (e.g., BNCI2014001 drops from 77.86\% to 75.12\%), and ST$\rightarrow$S yields a similar degradation. This performance disparity suggests that the spatial stream functions as a robust spatial filter, effectively enhancing the signal-to-noise ratio by aggregating spatial structural information to guide temporal feature extraction. Conversely, feeding complex, high-frequency temporal dynamics back into the spatial stream likely introduces stochastic noise, which disrupts the learning of stable spatial patterns due to the inherent variability of EEG signals. Thus, the asymmetric ST$\rightarrow$T flow, which prioritizes spatially-guided temporal refinement, remains the most effective strategy.

\subsection{Effect of Depth and Width}
To determine optimal hyperparameters, we evaluated the network depth and embedding width on the SEED and FACED datasets under the LOSO protocol. The depth configuration is denoted by $(N_s, N_t)$, where $N_s$ and $N_t$ represent the number of blocks in the spatial and temporal streams, respectively. We constrained the depth search space to $N_s \le N_t$ because the asymmetric $ST \rightarrow T$ mechanism relies on spatial features guiding temporal layers. Consequently, setting $N_s > N_t$ is invalid as the surplus spatial blocks cannot participate in the interaction. As shown in Figure \ref{fig:depth_width}(a) and (c), performance peaks at the $(3,3)$ configuration before declining due to overfitting induced by excessive model complexity. Regarding embedding width, Figures \ref{fig:depth_width}(b) and (d) indicate that accuracy maximizes at 40. Beyond this threshold, performance degrades as excessive dimensionality introduces redundancy and noise. Therefore, we adopt the $(3,3)$ configuration and a width of 40 for all subsequent experiments.

\begin{figure}
    \centering
    \includegraphics[width=1\linewidth]{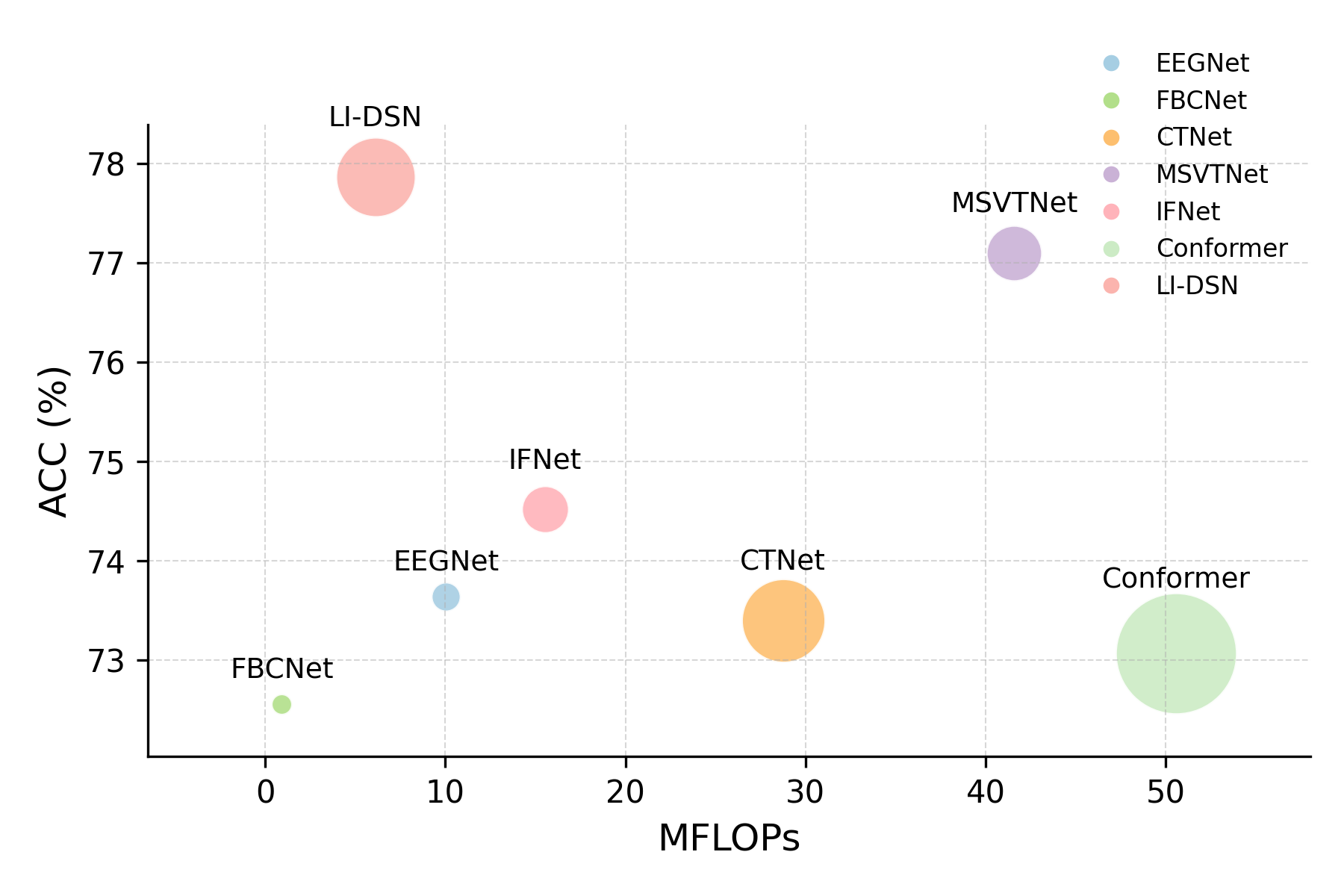}
    \caption{Comparison of model performance, parameters, and MFLOPS on BNCI2014001, with model size indicated by circle area.}
    \label{fig:cc}
\end{figure}

\subsection{Computational Complexity}
To verify the efficiency of LI-DSN, we analyzed its computational complexity in terms of parameters and Floating Point Operations (FLOPs) relative to baselines.  As illustrated in Figure \ref{fig:cc}, LI-DSN achieves a superior trade-off between performance and cost, locating at the optimal position with high accuracy and low resource consumption. Specifically, our model is extremely lightweight, requiring only 121.3 K parameters and 6.12 M FLOPs.  This represents a significant reduction in computational burden compared to heavy-weight deep learning models, while still maintaining the highest classification accuracy. This efficiency stems from the streamlined design of the TSIA mechanism, which effectively integrates temporal and spatial features without incurring the high computational overhead typical of complex attention architectures, making LI-DSN highly suitable for real-time BCI applications on resource-constrained devices.

\begin{figure*}
    \centering
    \makebox[\linewidth][r]{
       \includegraphics[width=1\linewidth]{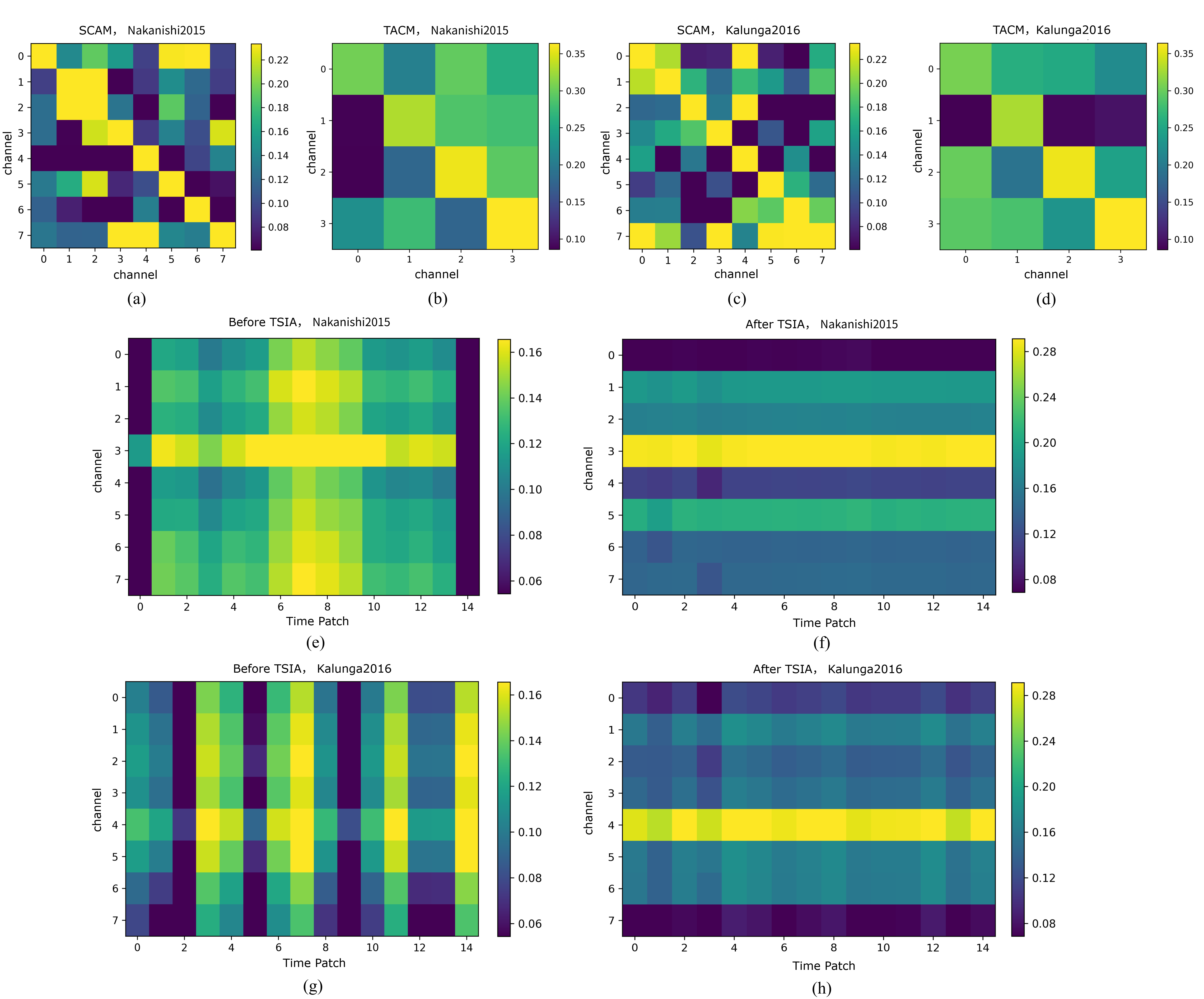}}
    \caption{Visualization of the TSIA matrices and channel-temporal interactions. Subfigures (a), (b), (e), and (f) correspond to the Nakanishi2015 dataset, while (c), (d), (g), and (h) correspond to the Kalunga2016 dataset. Specifically, (a) and (c) illustrate the spatial attention heatmap of SACM; (b) and (d) display the spatial attention heatmap of TCAM. Regarding the feature interaction, (e) and (g) depict the channel-temporal feature matrices before TSIA interaction, whereas (f) and (h) visualize the  matrices after interaction.}
    \label{fig:ah}
\end{figure*}

\subsection{Interpretability and Visualization}

\subsubsection{Learned Attention Patterns of SACM and TCAM}
To validate the effectiveness of the TSIA mechanism, we visualize the learned attention patterns on the Nakanishi2015 and Kalunga2016 datasets, as shown in the top row of Figure \ref{fig:ah}. The SACM in subfigures (a) and (c) exhibit distinct and sparse activation patterns rather than a uniform noise distribution. This indicates that the model successfully captures robust functional connectivity among channels, implicitly identifying salient spatial dependencies while suppressing redundant interactions. The TCAM in subfigures (b) and (d) display structured, block-wise attention distributions. This non-uniformity confirms that the network utilizes the cosine-gated mechanism to dynamically modulate temporal features, selectively emphasizing discriminative temporal features critical for decoding.

\subsubsection{Feature Evolution via TSIA Interaction}
Crucially, comparing the feature maps before and after fusion (bottom row of Figure \ref{fig:ah}) reveals the transformative impact of spatial guidance. The pre-interaction maps (e, g) exhibit disordered vertical fluctuations, suggesting that the initial representations lack consistent spatial structure and are susceptible to temporal instability. In sharp contrast, the post-interaction maps (f, h) evolve into distinct horizontal stratification. As observed, specific task-relevant channels exhibit consistent high activation across time windows while irrelevant background activities are effectively suppressed. This transition from vertical irregularity to horizontal stability demonstrates that the asymmetric design functions as an effective spatiotemporal filter, leveraging stable spatial priors to suppress background noise and lock onto task-relevant signal components.

\begin{figure}
    \centering
    \includegraphics[width=1\linewidth]{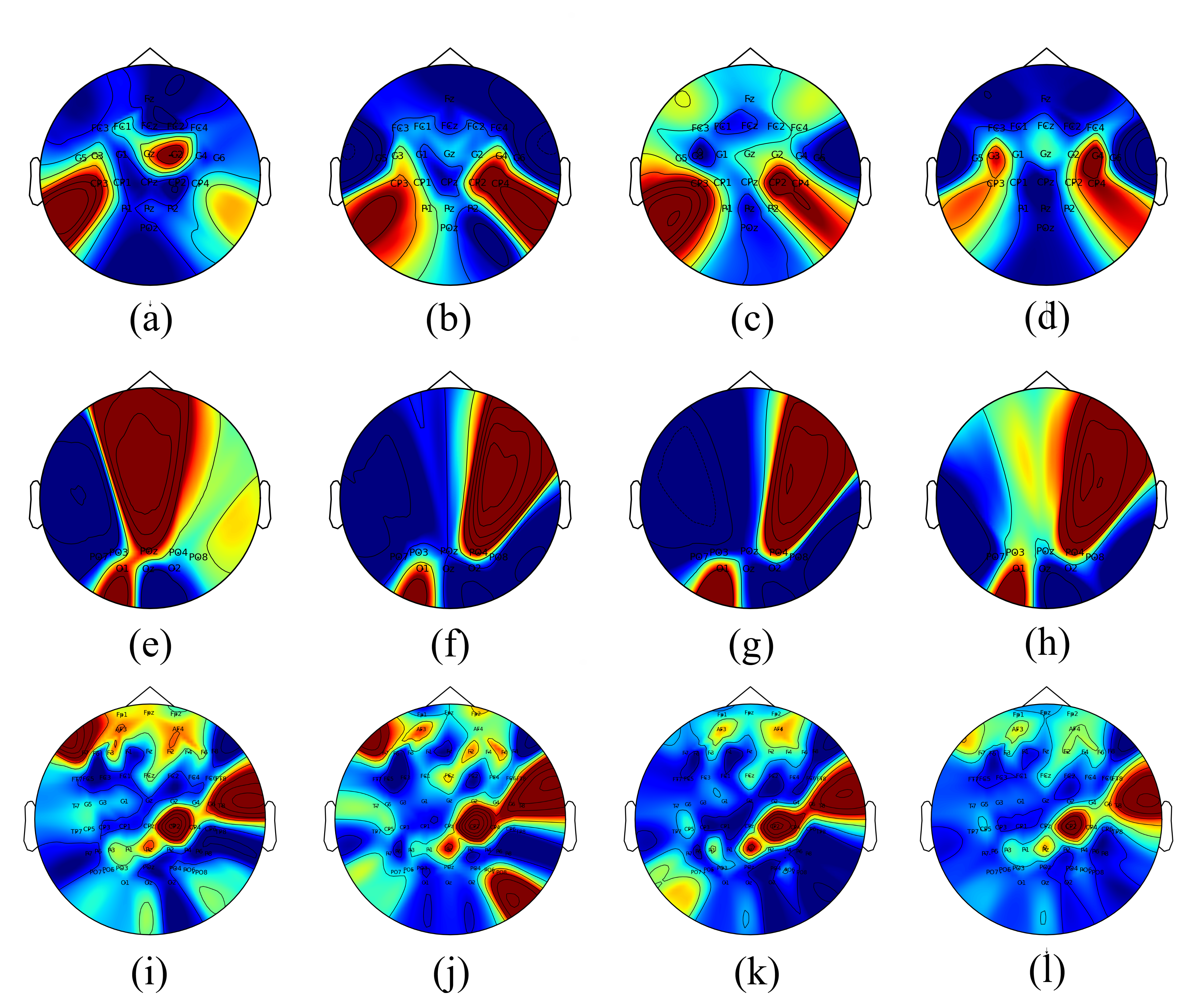}
    \caption{Visualization of saliency maps across three representative datasets. Subfigures (a)-(d) correspond to BNCI2014001, (e)-(h) represent the Kalunga2016 dataset, and (i)-(l) depict the SEED dataset. Within each group, the first three subfigures display activation patterns of representative subjects to illustrate individual variability, while the last subfigure presents the mean saliency map averaged across all subjects in the respective dataset.}
    \label{fig:saliencymaps}
\end{figure}

\subsubsection{Saliency Maps Visualization}
To validate the neurophysiological plausibility of LI-DSN, we selected three representative datasets corresponding to three distinct downstream EEG tasks: BNCI2014001, Kalunga2016, and SEED. As shown in Figure \ref{fig:saliencymaps}, we display activation patterns from three representative subjects alongside the group-level average for each dataset to illustrate both inter-subject variability and common neural signatures. The results demonstrate distinct, task-specific activation patterns consistent with established neuroscience findings. For MI classification (Figures \ref{fig:saliencymaps}(a)-(d)), high activations in the average map are concentrated in the sensorimotor cortex (e.g., C3, C4), aligning with the lateralized ERD/ERS phenomena \cite{pfurtscheller2001motor}. For SSVEP tasks (Figures \ref{fig:saliencymaps}(e)-(h)), saliency is predominantly localized in the posterior brain regions, specifically the occipital and parieto-occipital areas (e.g., O1, PO). This localization reflects the robust response of the visual cortex to the periodic flickering stimuli \cite{wang2006practical}. Conversely, in Emotion Recognition (Figures \ref{fig:saliencymaps}(i)-(l)), the model consistently highlights the right temporal (e.g., T8, FT8), central-parietal (e.g., CP2), and midline parietal (e.g., Pz) regions. These regions are part of the broader prefrontal and temporal lobes, which are critical for emotional regulation and processing \cite{zheng2015investigating}. These biologically consistent patterns confirm that LI-DSN effectively captures interpretable, task-relevant neural signatures despite individual differences.

\section{Limitations and Future Work}
Despite achieving superior performance, LI-DSN has several limitations that warrant future investigation. The current layer-wise interactive fusion mechanism, while effective, represents only one approach to integrating temporal and spatial information, and alternative interaction strategies such as attention-based cross-modal fusion or dynamic routing mechanisms may further enhance feature integration. The model's generalizability across different EEG paradigms beyond MI classification, emotion recognition and SSVEP requires further validation, and significant inter-subject variability remains challenging despite improved cross-subject performance. Future work should focus on exploring more sophisticated fusion architectures that adaptively weight stream contributions based on task demands and signal quality, investigating transfer learning frameworks to leverage knowledge across different BCI paradigms and reduce training data requirements, and conducting extensive clinical validation with neurological patients to establish practical utility in medical applications such as stroke rehabilitation and assistive communication systems.

\section{Conclusion}
This paper presents LI-DSN, a Layer-wise Interactive Dual-Stream Network for EEG decoding in brain-computer interfaces. By processing temporal dynamics and spatial relationships through separate streams with hierarchical interactive fusion via the TSIA mechanism, our approach effectively captures complementary spatiotemporal patterns in neural signals. Comprehensive experiments on MI classification, emotion recognition, and SSVEP datasets demonstrate that LI-DSN achieves superior performance compared to 13 baselines in both within-subject and cross-subject scenarios. The dual-stream architecture with layer-wise interactions enables effective integration of temporal and spatial information throughout the feature learning process, while ablation studies confirm the necessity of each component. As demonstrated through visualization analysis, the learned representations and the adaptive fusion capture meaningful temporal patterns and spatial connectivity, providing interpretable insights into neural decoding processes and offering a promising direction for developing more effective EEG-based brain-computer interface systems.

\bibliographystyle{IEEEtran} 
\bibliography{reference}

@inproceedings{wolpaw2007brain,
  title={{Brain-computer interfaces (BCIs) for communication and control}},
  author={{Wolpaw, Jonathan R}},
  booktitle={{Proceedings of the 9th international ACM SIGACCESS conference on Computers and accessibility}},
  pages={1--2},
  year={2007}
}

@article{moghimi2013review,
  title={{A review of EEG-based brain-computer interfaces as access pathways for individuals with severe disabilities}},
  author={{Moghimi, Saba and Kushki, Azadeh and Marie Guerguerian, Anne and Chau, Tom}},
  journal={{Assistive Technology}},
  volume={25},
  number={2},
  pages={99--110},
  year={2013},
  publisher={{Taylor \& Francis}}
}

@article{murad2024unveiling,
  title={{Unveiling thoughts: A review of advancements in EEG brain signal decoding into text}},
  author={{Murad, Saydul Akbar and Rahimi, Nick}},
  journal={{IEEE Transactions on Cognitive and Developmental Systems}},
  year={2024},
  publisher={{IEEE}}
}

@article{xie2022transformer,
  title={{A transformer-based approach combining deep learning network and spatial-temporal information for raw EEG classification}},
  author={{Xie, Jin and Zhang, Jie and Sun, Jiayao and Ma, Zheng and Qin, Liuni and Li, Guanglin and Zhou, Huihui and Zhan, Yang}},
  journal={{IEEE Transactions on Neural Systems and Rehabilitation Engineering}},
  volume={30},
  pages={2126--2136},
  year={2022},
  publisher={{IEEE}}
}

@article{salami2022eeg,
  title={{EEG-ITNet: An explainable inception temporal convolutional network for motor imagery classification}},
  author={{Salami, Abbas and Andreu-Perez, Javier and Gillmeister, Helge}},
  journal={{Ieee Access}},
  volume={10},
  pages={36672--36685},
  year={2022},
  publisher={{IEEE}}
}

@inproceedings{gong2023astdf,
  title={{ASTDF-net: attention-based spatial-temporal dual-stream fusion network for EEG-based emotion recognition}},
  author={{Gong, Peiliang and Jia, Ziyu and Wang, Pengpai and Zhou, Yueying and Zhang, Daoqiang}},
  booktitle={{Proceedings of the 31st ACM international conference on multimedia}},
  pages={883--892},
  year={2023}
}

@article{amin2021attention,
  title={{Attention-inception and long-short-term memory-based electroencephalography classification for motor imagery tasks in rehabilitation}},
  author={{Amin, Syed Umar and Altaheri, Hamdi and Muhammad, Ghulam and Abdul, Wadood and Alsulaiman, Mansour}},
  journal={{IEEE Transactions on Industrial Informatics}},
  volume={18},
  number={8},
  pages={5412--5421},
  year={2021},
  publisher={{IEEE}}
}

@article{zhao2024ctnet,
  title={{CTNet: a convolutional transformer network for EEG-based motor imagery classification}},
  author={{Zhao, Wei and Jiang, Xiaolu and Zhang, Baocan and Xiao, Shixiao and Weng, Sujun}},
  journal={{Scientific reports}},
  volume={14},
  number={1},
  pages={20237},
  year={2024},
  publisher={{Nature Publishing Group UK London}}
}

@ARTICLE{wang2025dbconformer,
  author={Wang, Ziwei and Wang, Hongbin and Jia, Tianwang and He, Xingyi and Li, Siyang and Wu, Dongrui},
  journal={{IEEE Journal of Biomedical and Health Informatics}}, 
  title={{DBConformer: Dual-Branch Convolutional Transformer for EEG Decoding}}, 
  year={2025},
  volume={},
  number={},
  pages={1-14},
  doi={10.1109/JBHI.2025.3622725}}

@article{jiruska2013synchronization,
  title={{Synchronization and desynchronization in epilepsy: controversies and hypotheses}},
  author={{Jiruska, Premysl and De Curtis, Marco and Jefferys, John GR and Schevon, Catherine A and Schiff, Steven J and Schindler, Kaspar}},
  journal={{The Journal of physiology}},
  volume={591},
  number={4},
  pages={787--797},
  year={2013},
  publisher={{Wiley Online Library}}
}

@article{schirrmeister2017deep,
  title={{Deep learning with convolutional neural networks for EEG decoding and visualization}},
  author={{Schirrmeister, Robin Tibor and Springenberg, Jost Tobias and Fiederer, Luk{\'a}{\v{s}} Josef and Glasstetter, Martin and Eggensperger, Katharina and Tangermann, Michael and Hutter, Frank and Burgard, Wolfram and Ball, Tonio}},
  journal={{Human brain mapping}},
  volume={38},
  number={11},
  pages={5391--5420},
  year={2017},
  publisher={{Wiley Online Library}}
}

@article{mane2021fbcnet,
  title={{FBCNet: A multi-view convolutional neural network for brain-computer interface}},
  author={{Mane, Ravikiran and Chew, Effie and Chua, Karen and Ang, Kai Keng and Robinson, Neethu and Vinod, A Prasad and Lee, Seong-Whan and Guan, Cuntai}},
  journal={{arXiv preprint arXiv:2104.01233}},
  year={2021}
}

@article{sakhavi2018learning,
  title={{Learning temporal information for brain-computer interface using convolutional neural networks}},
  author={{Sakhavi, Siavash and Guan, Cuntai and Yan, Shuicheng}},
  journal={{IEEE Transactions on Neural Networks and Learning Systems}},
  volume={29},
  number={11},
  pages={5619--5629},
  year={2018},
  publisher={{IEEE}}
}

@article{kwon2020subject,
  title={{Subject-independent brain--computer interfaces based on deep convolutional neural networks}},
  author={{Kwon, O-Yeon and Lee, Min-Ho and Guan, Cuntai and Lee, Seong-Whan}},
  journal={{IEEE Transactions on Neural Networks and Learning Systems}},
  volume={31},
  number={10},
  pages={3839--3852},
  year={2020},
  publisher={{IEEE}}
}

@article{cui2023eeg,
  title={{EEG-based cross-subject driver drowsiness recognition with an interpretable convolutional neural network}},
  author={{Cui, Jian and Lan, Zirui and Sourina, Olga and M{\"u}ller-Wittig, Wolfgang}},
  journal={{IEEE Transactions on Neural Networks and Learning Systems}},
  volume={34},
  number={10},
  pages={7921--7933},
  year={2023},
  publisher={{IEEE}}
}

@article{thuwajit2021eegwavenet,
  title={{EEGWaveNet: Multiscale CNN-based spatiotemporal feature extraction for EEG seizure detection}},
  author={{Thuwajit, Punnawish and Rangpong, Phurin and Sawangjai, Phattarapong and Autthasan, Phairot and Chaisaen, Rattanaphon and Banluesombatkul, Nannapas and Boonchit, Puttaranun and Tatsaringkansakul, Nattasate and Sudhawiyangkul, Thapanun and Wilaiprasitporn, Theerawit}},
  journal={{IEEE transactions on industrial informatics}},
  volume={18},
  number={8},
  pages={5547--5557},
  year={2021},
  publisher={IEEE}
}

@article{li2021novel,
  title={{A novel bi-hemispheric discrepancy model for EEG emotion recognition}},
  author={{Li, Yi and Wang, Lei and Zheng, Weilong and Zong, Yuan and Qi, Lei and Cui, Zhen and Zhang, Tong and Lu, Bao-Liang}},
  journal={{IEEE Transactions on Neural Networks and Learning Systems}},
  volume={32},
  number={8},
  pages={3477--3491},
  year={2021},
  publisher={{IEEE}}
}

@article{li2022ts,
  title={{TS-SEFFNet: Two-stage spatial-temporal feature fusion network for multisource EEG emotion recognition}},
  author={{Li, Yang and Wang, Lei and Zheng, Weilong and Zong, Yuan and Lu, Bao-Liang}},
  journal={{IEEE Transactions on Neural Networks and Learning Systems}},
  volume={33},
  number={10},
  pages={5629--5641},
  year={2022},
  publisher={{IEEE}}
}

@article{li2025neuron,
  title={{Neuron perception inspired EEG emotion recognition with parallel contrastive learning}},
  author={{Li, Dongdong and Huang, Shengyao and Xie, Li and Wang, Zhe and Xu, Jiazhen}},
  journal={{IEEE Transactions on Neural Networks and Learning Systems}},
  year={2025},
  publisher={{IEEE}}
}

@article{wang2025mvcnet,
  title={{MVCNet: Multi-view contrastive network for motor imagery classification}},
  author={{Wang, Z and Li, S and Chen, X and Wu, D}},
  journal={{Knowledge-Based Systems}},
  pages={114205},
  year={2025},
  publisher={Elsevier}
}

@article{sun2022dbgcn,
  author={{Sun, Mingyi and Cui, Weigang and Yu, Shuyue and Han, Hongbin and Hu, Bin and Li, Yang}},
  journal={{IEEE Transactions on Affective Computing}}, 
  title={{A Dual-Branch Dynamic Graph Convolution Based Adaptive TransFormer Feature Fusion Network for EEG Emotion Recognition}}, 
  year={2022},
  volume={13},
  number={4},
  pages={2218-2228},
  doi={{10.1109/TAFFC.2022.3199075}}
}

@ARTICLE{Li2025Dual-TSST,
  author={{Li, Hongqi and Zhang, Haodong and Chen, Yitong}},
  journal={{IEEE Journal of Biomedical and Health Informatics}}, 
  title={{Dual-TSST: A Dual-Branch Temporal-Spectral-Spatial Transformer Model for EEG Decoding}}, 
  year={2025},
  volume={29},
  number={9},
  pages={6524-6537},
  doi={{10.1109/JBHI.2025.3577611}}
}

@ARTICLE{Ye2025DS-AGC,
  author={{Ye, Weishan and Zhang, Zhiguo and Teng, Fei and Zhang, Min and Wang, Jianhong and Ni, Dong and Li, Fali and Xu, Peng and Liang, Zhen}},
  journal={{IEEE Transactions on Affective Computing}}, 
  title={{Semi-Supervised Dual-Stream Self-Attentive Adversarial Graph Contrastive Learning for Cross-Subject EEG-Based Emotion Recognition}}, 
  year={2025},
  volume={16},
  number={1},
  pages={290-305},
  doi={{10.1109/TAFFC.2024.3433470}}}

@ARTICLE{Huang2024FBSTCNet,
  author={{Huang, Weichen and Wang, Wenlong and Li, Yuanqing and Wu, Wei}},
  journal={{IEEE Transactions on Affective Computing}}, 
  title={{FBSTCNet: A Spatio-Temporal Convolutional Network Integrating Power and Connectivity Features for EEG-Based Emotion Decoding}}, 
  year={2024},
  volume={15},
  number={4},
  pages={1906-1918},
  doi={{10.1109/TAFFC.2024.3385651}}}

@inproceedings{liu2025review,
  title={{A review of multimodal medical data fusion techniques for personalized medicine}},
  author={Liu, Congcong and Ye, Fangfang},
  booktitle={{Proceedings of the 4th International Conference on Biomedical and Intelligent Systems}},
  pages={338--347},
  year={2025}
}

@ARTICLE{Gong2024Bi-Hemi,
  author={{Gong, Linlin and Chen, Wanzhong and Zhang, Dingguo}},
  journal={{IEEE Journal of Biomedical and Health Informatics}}, 
  title={{An Attention-Based Multi-Domain Bi-Hemisphere Discrepancy Feature Fusion Model for EEG Emotion Recognition}}, 
  year={2024},
  volume={28},
  number={10},
  pages={5890-5903},
  doi={{10.1109/JBHI.2024.3418010}}}

@article{bnci2014001,
  title={{Review of the BCI competition IV}},
  author={{Tangermann, M. and M{\"u}ller, K.-R. and Aertsen, A. and Birbaumer, N. and Braun, C. and Brunner, C. and Leeb, R. and Mehring, C. and Miller, K. J. and M{\"u}ller-Putz, G. R. and others}},
  journal={{Frontiers in Neuroscience}},
  volume={6},
  pages={55},
  year={2012},
}

@article{bnci2014002,
  title={{Random forests in non-invasive sensorimotor rhythm brain-computer interfaces: A practical and convenient non-linear classifier}},
  author={{Steyrl, D. and Scherer, R. and Faller, J. and M{\"u}ller-Putz, G. R.}},
  journal={{Biomedical Engineering/Biomedizinische Technik}},
  volume={61},
  number={1},
  pages={77--86},
  year={2016},
}

@article{bnci2014004,
  title={{Brain--computer communication: motivation, aim, and impact of exploring a virtual apartment}},
  author={{Leeb, R. and Lee, F. and Keinrath, C. and Scherer, R. and Bischof, H. and Pfurtscheller, G.}},
  journal={{IEEE Transactions on Neural Systems and Rehabilitation Engineering}},
  volume={15},
  number={4},
  pages={473--482},
  year={2007},
}

@article{Zhou2016,
  title={{A fully automated trial selection method for optimization of motor imagery based braincomputer interface}},
  author={{Zhou, B. and Wu, X. and Lv, Z. and Zhang, L. and Guo, X.}},
  journal={{PloS One}},
  volume={11},
  number={9},
  pages={e0162657},
  year={2016},
}

@article{seed,
  title={{Investigating critical frequency bands and channels for EEG-based emotion recognition with deep neural networks}},
  author={{Zheng, W.-L. and Lu, B.-L.}},
  journal={{IEEE Transactions on Autonomous Mental Development}},
  volume={7},
  number={3},
  pages={162--175},
  month={Sep},
  year={2015},
}

@article{faced,
  title={{A large finer-grained affective computing EEG dataset}},
  author={{Chen, J. and Wang, X. and Huang, C. and Hu, X. and Shen, X. and Zhang, D.}},
  journal={{Scientific Data}},
  volume={10},
  number={1},
  pages={740},
  month={Oct},
  year={2023},
}

@article{Nakanishi2015,
  title={{A comparison study of canonical correlation analysis based methods for detecting steady-state visual evoked potentials}},
  author={{Nakanishi, M. and Wang, Y. and Wang, Y. T. and Jung, T. P.}},
  journal={{PloS One}},
  volume={10},
  number={10},
  pages={1--18},
  year={2015},
}

@article{Kalunga2016,
  title={{Online SSVEP-based BCI using Riemannian Geometry}},
  author={{Kalunga, Emmanuel K. and Chevallier, Sylvain and Barthelemy, Quentin}},
  journal={Neurocomputing},
  year={2016},
}

@ARTICLE{Liu2025DMSACNN,
  author={{Liu, Ke and Xing, Xin and Yang, Tao and Yu, Zhuliang and Xiao, Bin and Wang, Guoyin and Wu, Wei}},
  journal={{IEEE Journal of Biomedical and Health Informatics}}, 
  title={{DMSACNN: Deep Multiscale Attentional Convolutional Neural Network for EEG-Based Motor Decoding}}, 
  year={2025},
  volume={29},
  number={7},
  pages={4884-4896},
  doi={{10.1109/JBHI.2025.3546288}}}

@article{lawhern2018eegnet,
  title={{EEGNet: a compact convolutional neural network for EEG-based brain--computer interfaces}},
  author={{Lawhern, Vernon J and Solon, Amelia J and Waytowich, Nicholas R and Gordon, Stephen M and Hung, Chou P and Lance, Brent J}},
  journal={{Journal of neural engineering}},
  volume={15},
  number={5},
  pages={056013},
  year={2018},
  publisher={iOP Publishing}
}

@article{zhang2019gcb,
  title={{GCB-Net: Graph Convolutional Broad Network and its Application in Emotion Recognition}},
  author={{Zhang, Tong and Wang, Xingliang and Xu, Xiangmin and Chen, C L Philip}},
  journal={IEEE Transactions on Affective Computing},
  volume={13},
  number={1},
  pages={379--388},
  year={2019},
  publisher={IEEE}
}

@article{liu2024msvtnet,
  title={{MSVTNet: Multi-scale vision transformer neural network for EEG-based motor imagery decoding}},
  author={{Liu, Ke and Yang, Tao and Yu, Zhuliang and Yi, Weibo and Yu, Hong and Wang, Guoyin and Wu, Wei}},
  journal={{IEEE Journal of Biomedical and Health Informatics}},
  year={2024},
  publisher={IEEE}
}

@article{wang2023ifnet,
  title={{IFNet: An Interactive Frequency Convolutional Neural Network for Enhancing Motor Imagery Decoding From EEG}},
  author={Wang, Jiaheng and Yao, Lin and Wang, Yu},
  journal={{IEEE Transactions on Neural Systems and Rehabilitation Engineering}},
  volume={31},
  pages={1900--1911},
  year={2023},
  publisher={IEEE}
}

@article{song2023eeg,
  title={{EEG Conformer: Convolutional Transformer for EEG Decoding and Visualization}},
  author={Song, Yonghao and Zheng, Qingqing and Liu, Bingchuan and Gao, Xiaorong},
  journal={{IEEE Transactions on Neural Systems and Rehabilitation Engineering}},
  volume={31},
  pages={710--719},
  year={2023},
  publisher={IEEE}
}

@article{ding2024eeg,
  title={{EEG-Deformer: A dense convolutional transformer for brain-computer interfaces}},
  author={Ding, Yi and Li, Yong and Sun, Hao and Liu, Rui and Tong, Chengxuan and Liu, Chenyu and Zhou, Xinliang and Guan, Cuntai},
  journal={{IEEE Journal of Biomedical and Health Informatics}},
  year={2024},
  publisher={IEEE}
}

@article{ding2025emt,
  title={{EmT: A novel transformer for generalized cross-subject EEG emotion recognition}},
  author={Ding, Yi and Tong, Chengxuan and Zhang, Shuailei and Jiang, Muyun and Li, Yong and Lim, Kevin JunLiang and Guan, Cuntai},
  journal={{IEEE Transactions on Neural Networks and Learning Systems}},
  year={2025},
  publisher={IEEE}
}

@article{ding2022tsception,
  title={{TSception: Capturing temporal dynamics and spatial asymmetry from EEG for emotion recognition}},
  author={Ding, Yi and Robinson, Neethu and Zhang, Su and Zeng, Qiuhao and Guan, Cuntai},
  journal={{IEEE Transactions on Affective Computing}},
  volume={14},
  number={3},
  pages={2238--2250},
  year={2022},
  publisher={IEEE}
}

@article{pfurtscheller2001motor,
  title={{Motor imagery and direct brain-computer communication}},
  author={Pfurtscheller, Gert and Neuper, Christa},
  journal={{Proceedings of the IEEE}},
  volume={89},
  number={7},
  pages={1123--1134},
  year={2001},
  publisher={IEEE}
}

@article{zheng2015investigating,
  title={{Investigating critical frequency bands and channels for EEG-based emotion recognition with deep neural networks}},
  author={Zheng, Wei-Long and Lu, Bao-Liang},
  journal={{IEEE Transactions on autonomous mental development}},
  volume={7},
  number={3},
  pages={162--175},
  year={2015},
  publisher={IEEE}
}

@article{wang2006practical,
  title={{A practical VEP-based brain-computer interface}},
  author={Wang, Yijun and Wang, Ruiping and Gao, Xiaorong and Hong, Bo and Gao, Shangkai},
  journal={{IEEE Transactions on Neural Systems and Rehabilitation Engineering}},
  volume={14},
  number={2},
  pages={234--240},
  year={2006},
}

\end{document}